\documentclass[10pt,twocolumn,letterpaper]{article}

\usepackage[pagenumbers]{iccv} 

\definecolor{iccvblue}{rgb}{0.21,0.49,0.74}
\usepackage[pagebackref,breaklinks,colorlinks,allcolors=iccvblue]{hyperref}
\usepackage{algorithm}
\usepackage{algorithmic}
\usepackage[most]{tcolorbox}
\usepackage{dashrule}
\usepackage{float}
\usepackage{multirow}

\def\paperID{7077} 
\def\confName{ICCV}
\def\confYear{2025}

\title{Reverse Prompt: Cracking the Recipe Inside Text-to-Image Generation}

\author{
Zhiyao Ren$^{1}$,
Yibing Zhan$^{2}$,
Baosheng Yu$^{1}$,
Dacheng Tao$^{1}$ \\
$^{1}$Nanyang Technological University, Singapore\\ $^{2}$JD Explore Academy, China \\
\texttt{zhiyao001@e.ntu.edu.sg}, \texttt{zhanyibing@jd.com}, \\ \texttt{baosheng.yu@ntu.edu.sg}, \texttt{dacheng.tao@gmail.com}
}

\begin{document}
\maketitle
\begin{abstract}

Text-to-image generation has become increasingly popular, but achieving the desired images often requires extensive prompt engineering. In this paper, we explore how to decode textual prompts from reference images, a process we refer to as \textbf{image reverse prompt engineering}. This technique enables us to gain insights from reference images, understand the creative processes of great artists, and generate impressive new images. To address this challenge, we propose a method known as \textbf{automatic reverse prompt optimization} (ARPO). 
Specifically, our method refines an initial prompt into a high-quality prompt through an iteratively imitative gradient prompt optimization process: 1) generating a recreated image from the current prompt to instantiate its guidance capability; 2) producing textual gradients, which are candidate prompts intended to reduce the difference between the recreated image and the reference image; 3) updating the current prompt with textual gradients using a greedy search method to maximize the CLIP similarity between prompt and reference image.
We compare ARPO with several baseline methods, 
including handcrafted techniques, gradient-based prompt tuning methods, image captioning, and data-driven selection method. 
Both quantitative and qualitative results demonstrate that our ARPO converges quickly to generate high-quality reverse prompts. More importantly, we can easily create novel images with diverse styles and content by directly editing these reverse prompts. Code will be made publicly available.
\end{abstract}

\section{Introduction}

In recent years, Artificial Intelligence Generated Content (AIGC) has gained significant popularity, largely driven by the success of text-driven generation applications such as text-to-image~\cite{rombach2022ldm,podell2023sdxl,saharia2022imagen,betker2023dalle3,chen2023pixart}, text-to-video~\cite{ho2022video,blattmann2023vldm,singer2022makeavideo}, and text-to-3D~\cite{lee2024dreamflow, chen2024vp3d}. Meanwhile, there has been a significant increase in demand for production-ready text prompts, known as prompt engineering. This process often requires considerable effort in meticulous crafting and iteration to ensure precision, clarity, and optimization for the intended use case.
AIGC communities have provided many demo images with prompt templates\footnote{https://lexica.art;~https://civitai.com;~https://stablediffusionweb.com}, lowering the barrier to entry for AIGC. This accessibility encourages more individuals to create their desired images by modifying these templates. However, there is a growing need for a more scalable and automated approach to explore rich textual prompts. 

\begin{figure}[!t]
    \centering
    \includegraphics[width=\linewidth]{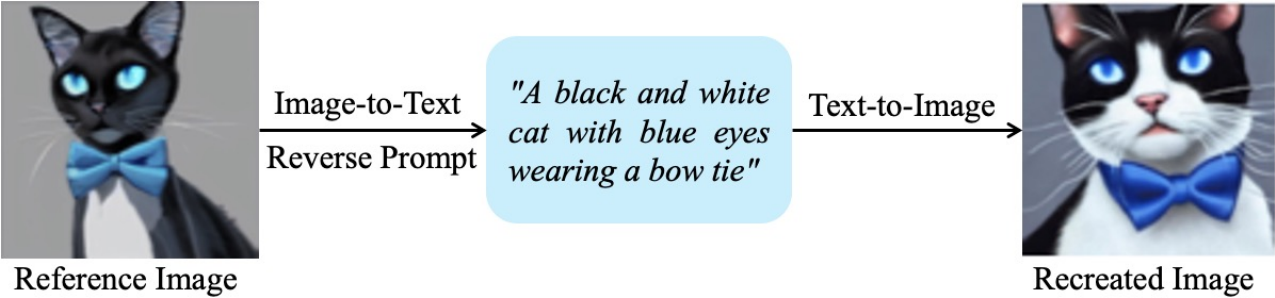}
    \caption{Illustration of image reverse prompt engineering. Given a reference image, the goal is to identify a reverse prompt that effectively recreates the image with similar content and style.}
    \label{fig:reverse-prompt-example}
\end{figure}

\begin{table*}[t]
    \centering
    \small
    \caption{Illustration of reverse prompts using different methods.}
    \label{tab:showcase}
    \setlength\tabcolsep{3pt}
    \scalebox{0.85}{
    \begin{tabular}{|c|c|l|}
         \hline
         Method & Characterization & \multicolumn{1}{c|}{Reverse Prompt}  \\ \hline
        Hand-Crafted & \begin{tabular}[c]{@{}c@{}}Brief,\\ Inaccurate\end{tabular} & \begin{tabular}[c]{@{}l@{}}fantastical landscape, digital painting style, huge landscape, people in the middle of the road, green trees\end{tabular}  \\ \hline
        PH2P & Poor Readability & \begin{tabular}[c]{@{}l@{}} the pic mew amazing god th A devils awe down the astronomy lakel swallowed rooftop !! , , walking\\ ... barrabest A flooded road compromising painting by A  walking into A dirt road falsely painting \end{tabular}\\ \hline
        GPT4V & \begin{tabular}[c]{@{}c@{}}Useless Information, \\Unrecognisable\end{tabular} & \begin{tabular}[c]{@{}l@{}}The scene features futuristic structures, lush greenery, and a dramatic sky with swirling clouds and\\ ... elements of fantasy and science fiction, evoking a sense of wonder and exploration.\end{tabular}  \\ \hline
        CLIP-Interrogator & Restricted to dataset &\begin{tabular}[c]{@{}l@{}}brightly lit building with a bench in front of it, rendering of log pile factory, colorful architectural\\ ... inspired by Hugh Ferriss, with aurora borealis in the sky, surreal sci fi architecture\end{tabular}  \\ \hline
        ARPO (ours) & \begin{tabular}[c]{@{}c@{}}Readability,\\Recognisable,\\ Dataset-independent\end{tabular} & \begin{tabular}[c]{@{}l@{}}imaginative landscape, surreal quality, exaggerated proportions, blend of realism and fantasy, dramatic sky,\\ exploration,figures placed centrally, digital painting style with high level of detail and sense of depth\end{tabular} \\ \hline
    \end{tabular}
    }
\end{table*}

\begin{figure*}[!ht]
\centering
\begin{subfigure}{0.145\linewidth}\includegraphics[width=1\textwidth]{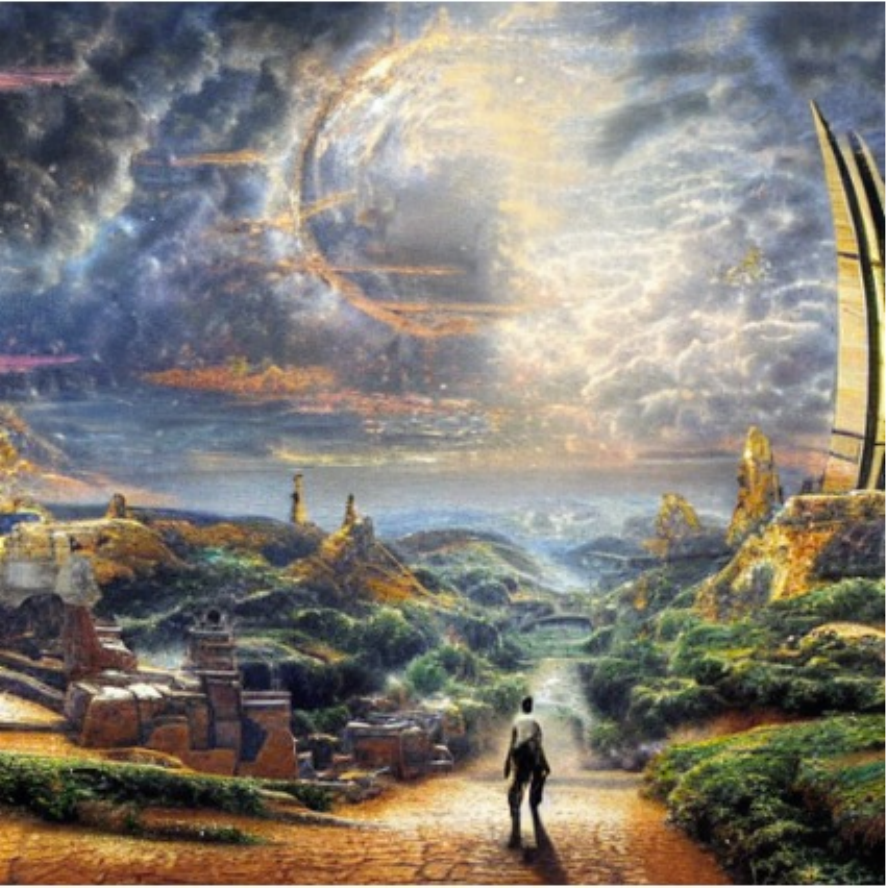}
\caption{Reference Image}
\end{subfigure}\quad
\begin{subfigure}{0.145\linewidth}\includegraphics[width=1\textwidth]{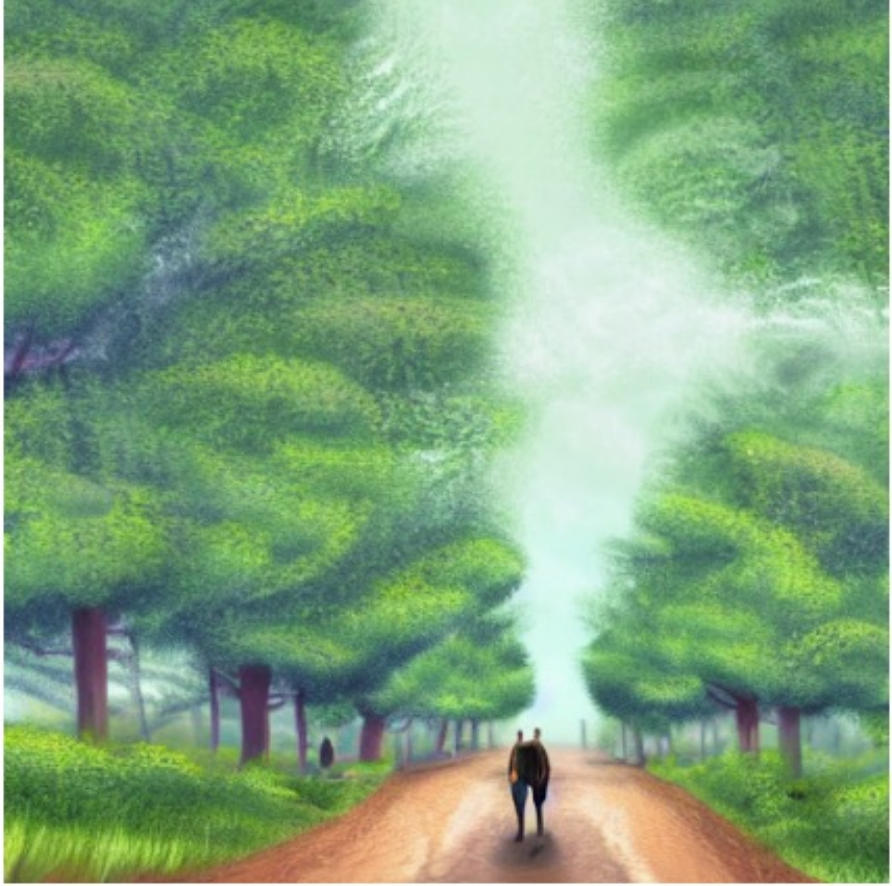}
\caption{Hand-Crafted}
\end{subfigure}\quad
\begin{subfigure}{0.145\linewidth}\includegraphics[width=1\textwidth]{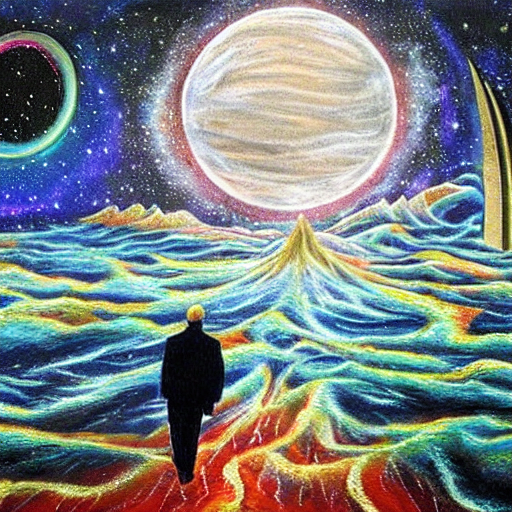}
\caption{PH2P}
\end{subfigure}\quad
\begin{subfigure}{0.145\linewidth}\includegraphics[width=1\textwidth]{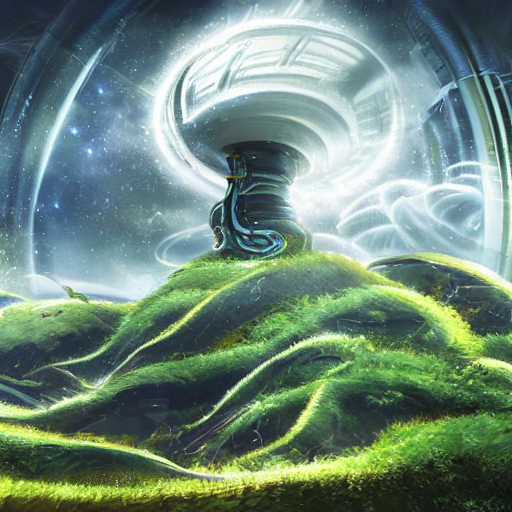}
\caption{GPT-4V}
\end{subfigure}\quad
\begin{subfigure}{0.145\linewidth}\includegraphics[width=1\textwidth]{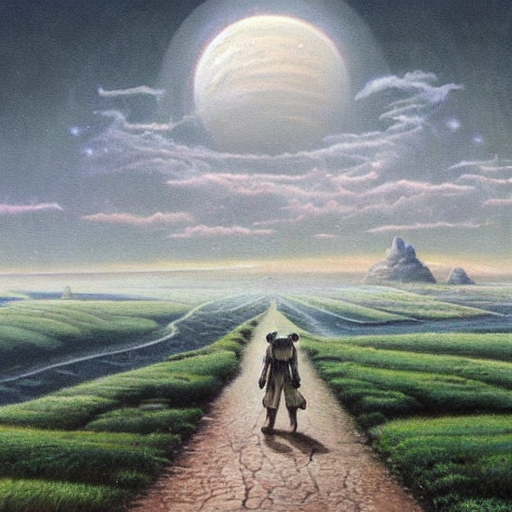}
\caption{CLIP-Interrogator}
\end{subfigure}\quad
\begin{subfigure}{0.145\linewidth}\includegraphics[width=1\textwidth]{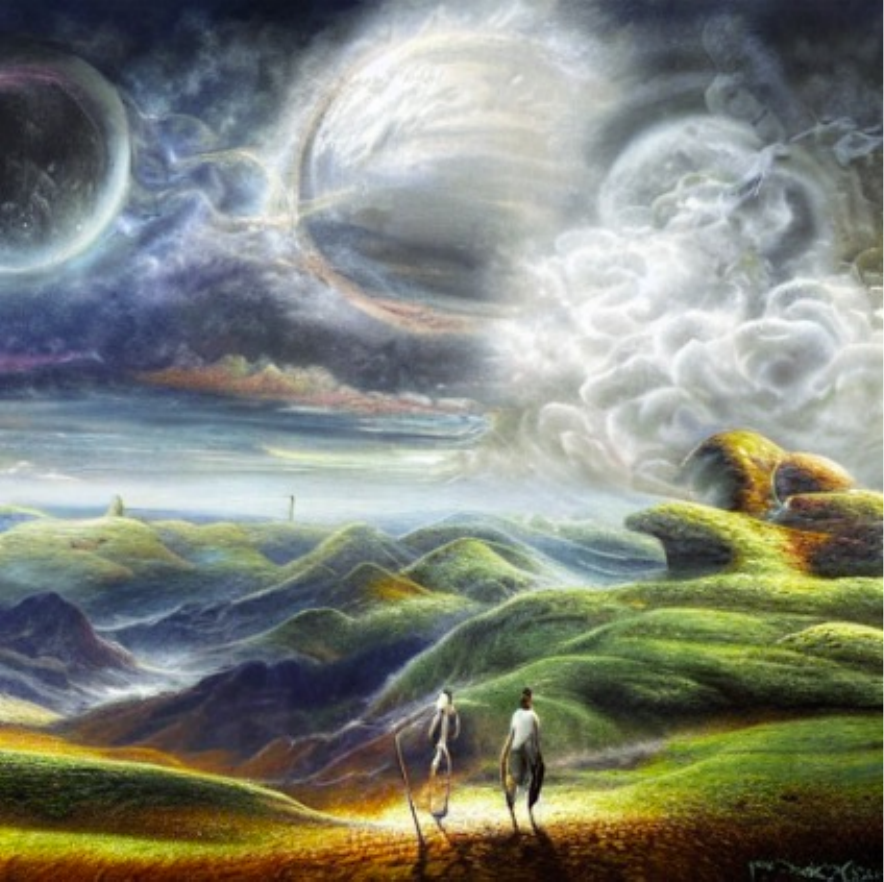}
\caption{APRO (ours)}
\end{subfigure}
\caption{Illustration of recreated images using different reverse prompts. 
}
\label{fig:showcase}
\end{figure*}

Figure~\ref{fig:reverse-prompt-example} illustrates a promising approach: crafting a prompt capable of generating a given reference image, which we refer to as image reverse prompt engineering (IRPE). This process not only uncovers the relationship between textual prompts and reference images but also provides a good starting point for generating novel images by modifying the reverse prompts derived from reference images. Notably, this paper considers the reverse prompt by focusing on images that share similar content and style at the semantic level, rather than on reconstructing pixel-level details. Recently, there have been several attempts at IRPE, but these methods still exhibit shortcomings. Gradient-based methods such as PEZ~\cite{wen2024hard} and PH2P~\cite{mahajan2024prompting} have two main drawbacks: 1) they are not compatible with black-box models or API services; and 2) the reverse prompts generated by directly converting from a continuous space often result in unclear expressions. As shown in Table~\ref{tab:showcase}, the reverse prompt generated by PH2P is not in natural language and is difficult to understand. An alternative solution is to utilize the image captioning capabilities of vision-language models (VLMs), 
such as BLIP~\cite{li2022blip,li2023blip2,dai2024instructblip}, LLaVa~\citep{liu2024llava,liu2023llava1.5,liu2024llava-next}, VILA~\citep{lin2023vila}, and GPT-4V~\citep{achiam2023gpt}. 
However, these methods may include information that cannot be recognized by the generative model. For example, as shown in Table~\ref{tab:showcase}, the prompt describes ``\textit{A solitary figure is walking down a worn.}", but this detail is not included in the recreated image. Additional, CLIP-Interrogator~\cite{clip-interrogator}
rely on large-scale datasets to capture specific characteristics; however, as shown in the Table~\ref{tab:showcase}, it struggles to accurately describe the style of the image due to the limitations in the comprehensiveness of the dataset. Moreover, some services do not provide sufficient technical details\footnote{https:www.phot.ai;~ https://animegenius.live3d.io/}. Figure~\ref{fig:showcase} illustrates images generated from different reverse prompts of Table~\ref{tab:showcase}, clearly highlighting the noticeable differences between the recreated and reference images.

To address the challenge of image reverse prompt engineering, we propose an automatic reverse prompt optimization (ARPO) framework. Given a reference image, we first initialize the reverse prompt using VLMs such as BLIP2, or we can also start with a hand-crafted prompt. 
We then iteratively optimize the prompt through three key steps: 1) \textbf{Image Generation}: A text-to-image model recreates an image based on the current prompt, demonstrating its descriptive capability and its adaptability to the generative model; 2) \textbf{Prompt Generation}: A prompt generator, powered by LLMs/VLMs, produces candidate prompts as textual gradients to reduce the difference between the recreated image and reference image. 
Inspired by the chain-of-thought (COT) strategy \cite{wei2022chain}, we utilize a two-step framework that first describing differences between the recreated image and the reference image and then generating prompts, eliciting the capability of LLMs/VLMs. For open-source VLMs with lower performing, we further propose a more refined enhanced framework; and 3) \textbf{Prompt Selection}: A greedy algorithm identifies the optimal candidate prompts for updating the current prompt, ensuring that the prompts are optimized to maximize the similarity between reverse prompt and image. 
For the simplicity, we use CLIP-score as the similarity.
Our ARPO offers several advantages over existing methods: 1) Unlike gradient-based approaches, we use a gradient imitation strategy~\cite{pryzant2023automatic} for textual prompts, making it compatible with black-box models and API services, while allowing the reverse prompt to be viewed and edited as text; 2) Compared to image captioning methods, our approach is highly adaptable to specific text-to-image models; and 3) In contrast to data-driven methods such as CLIP-Interrogator, our ARPO framework does not require large-scale external training data.
Lastly, our goal is not only to recreate the reference image but, more importantly, to enable the generation of novel images with similar content and/or style to the reference image by editing the derived reverse prompt, as detailed in Section~\ref{sec:method:novel-image}.

\begin{figure*}[!t]
    \centering
    \includegraphics[width=0.75\linewidth]{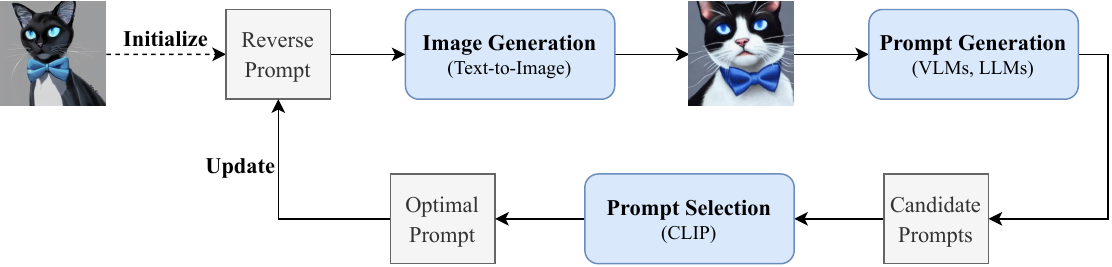}
    \caption{\textbf{The main ARPO framework} consists of three main components: image generation, prompt generation, and prompt selection.}
    \label{fig:framework}
    \vspace{-2mm}
\end{figure*}

We conduct both quantitative and qualitative comparisons between our ARPO framework and existing methods/services by introducing a new dataset that includes 50 AI-generated images \citep{wang2022diffusiondb} and 50 human-created images \citep{saleh2015wikiart,jing2020dynamic}. For the quantitative evaluation of prompt and image fidelity, we use four metrics: \texttt{CLIP-T}, \texttt{CLIP-I}, \texttt{ViT}, and \texttt{DINO}~\citep{radford2021clip,caron2021DINO,dosovitskiy2020vit}. In addition, we perform qualitative comparisons focusing on both image recreation and novel image generation applications, including image modification and fusion.  Extensive experimental results demonstrate the effectiveness of the proposed ARPO framework for image reverse prompt engineering, highlighting its potential in the AIGC domain. We will make all code and data publicly available to facilitate further research in this area.
Our main contributions can be summarized as follows:
\begin{itemize}
    \item We introduce a novel automatic reverse prompt optimization (ARPO) framework using gradient imitation strategy to generate high-quality reverse prompts, which can be applied to both black-box and white-box AIGC methods and modified to create novel images with content and/or styles similar to the reference image.
    \item We conduct comprehensive qualitative and quantitative experiments, illustrating the efficacy of our proposed method for image reverse prompt engineering.
    We showcase the exceptional results of our ARPO in image modification and fusion, highlighting the potential of reverse prompts for diverse creative tasks.
\end{itemize}

\section{Related Work}
\noindent\textbf{Image Reverse Prompt Engineering.}
Previous methods can be categorized into three main types: \textbf{1) Gradient-based methods} typically involve soft prompt tuning followed by the discretization, where optimized embeddings are re-projected onto the nearest neighbors in vocabulary. For example, \citet{wen2024hard} optimizes hard prompts using image-text similarity matching to generate reverse prompts, while \citet{williams2024prompt} applies various discrete prompt tuning algorithms in adversarial attacks, such as GCG \cite{zou2023GCG}, AutoDAN \cite{zhu2023autodan}, and Random Search \cite{andriushchenko2023randomsearch}. Furthermore, \citet{mahajan2024prompting} leverages a text-conditioned diffusion model to optimize prompts. However, these approaches rely on continuous embeddings, often producing reverse prompts with unreadable expressions and limited interpretability;
\textbf{2) Image captioning methods} utilize VLMs such as LLaVA \citep{liu2024llava,liu2023llava1.5,liu2024llava-next}, ShareGPT-4V \citep{chen2023sharegpt4v}, VILA \citep{lin2023vila}, and GPT-4V \citep{achiam2023gpt} to generate reverse prompts from reference images. While VLMs provide comprehensive image descriptions, their outputs tend to be lengthy and complex for guiding text-to-image generation; and 
\textbf{3) Data-driven methods}, such as CLIP-Interrogator~\citep{clip-interrogator}, require a large-scale prior dataset. This method first compiles a dataset of over 100,000 image descriptions in tag format, then uses cosine similarity between image and text embeddings to filter this dataset and create prompts. A significant limitation is that the generated prompts are restricted to the dataset used. In contrast to these methods, our ARPO generates reverse prompts through automatic prompt optimization, offering a more general and effective approach.

\noindent\textbf{Automatic Prompt Optimization.}
Creating high-quality prompts manually is time-consuming and labor-intensive, requiring significant expertise. Consequently, there is a growing demand for methods that automate prompt generation and optimization. These methods generally fall into two categories: real-gradient methods and imitated-gradient methods. Real-gradient methods optimize prompts by calculating numerical gradients for search or fine-tuning. For instance, AutoPrompt \citep{shin2020autoprompt} uses gradient-guided search to identify trigger tokens, which are then combined with the original task input to create prompts. Imitated-gradient methods, in contrast, replace numerical gradients with other approaches. 
APE~\cite{zhou2022large} obtain the optimize prompts by assigning LLMs to perform inference, scoring, and resampling while APO~\cite{pryzant2023automatic} introduces text gradients as a substitute for numerical gradients.
In the context of image generation, Idea2Img~\cite{yang2024idea2img} iteratively refines prompts to better align with user intent.
In contract to previous methods, our approach is the first to apply automatic prompt optimization to image reverse prompt engineering.

\begin{figure*}[!t]
    \centering
    \includegraphics[width=0.75\linewidth]{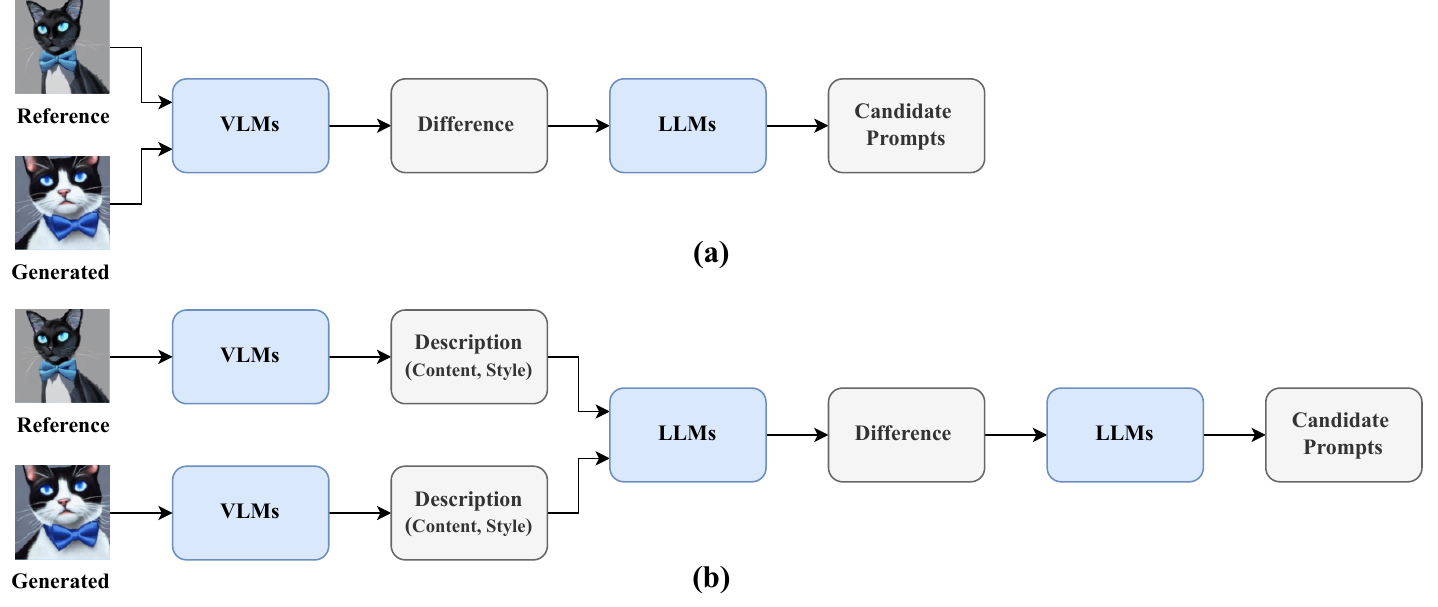}
    \caption{\textbf{Two prompt generation frameworks used in ARPO.} (a) Vanilla prompt generation. (b) Enhanced prompt generation.}
    \label{fig:prompt-generation}
    \vspace{-2mm}
\end{figure*}

\section{Method}
\label{sec:method}

In this section, we first describe image reverse prompt engineering (IRPE). We then introduce the proposed ARPO framework. Lastly, we discuss novel image generation.

\subsection{Image Reverse Prompt Engineering}
\label{sec:method:irpe}


Given a reference image  $I_g$ and a text-to-image model $f$, we explore whether there exists a prompt $P$ such that $I_g = f(P)$. This raises the question of whether an inverse function $f^{-1}$  exists, allowing us to understand the image generation process: $P = f^{-1}(I_g)$. Directly learning such invertible generative models is inherently challenging, as shown in prior research~\cite{dinh2014nice,kingma2018glow,chen2019residual,kobyzev2020normalizing,behrmann2019invertible}. Additionally, exploring flow-based generative models or invertible networks extends beyond this paper's scope. Therefore, given that most popular text-driven generation models are non-invertible, we avoid assumptions about the text-to-image models themselves. Instead, we aim to offer a practical and accessible solution through prompt engineering: we decode a reference image $I_r$  to derive a reverse prompt  $P$  that best maintains essential image attributes, such as content and style. This process can be formulated as 
\begin{equation}
    P^* \in \texttt{argmin}_{P}~\mathcal{L}(I_r, f(P)),
\end{equation}
where $\mathcal{L}$ represents the image discrepancy criterion.

\subsection{Automatic Reverse Prompt Optimization}
\label{sec:method:ARPO}

The main ARPO framework is shown in Figure~\ref{fig:framework}. Below, we detail the processes of prompt initialization, image generation, prompt generation, and prompt selection.

\noindent\textbf{Prompt Initialization.}
Recently, creating a simple reverse prompt has become very easy using either image caption models, VLMs, or hand-crafted approaches. We thus begin automatic prompt optimization with an initial reverse prompt to significantly reduce the required iterations. In this paper, we utilize the image caption interface of the popular BLIP2~\citep{li2023blip2} model for prompt initialization. For instance, the initialized simple reverse prompt for the reference image in Figure~\ref{fig:framework} is ``\textit{a black and white cat with blue eyes wearing a bow tie}”. That is, the initialized reverse prompt $P^0$ for a given reference image $I_r$ is:
$P^0 = \texttt{BLIP2}(I_r)$.

\noindent\textbf{Image Generation.} 
Since text-to-image models often exhibit distinct instruction tuning preferences, we do not aim for a model-agnostic process. Instead, we optimize the reverse prompt based on feedback from each specific text-to-image model: at the $t$-th iteration, we employ current reverse prompt $P^t$ to generate a new image $I_g^t$ as follows:
\begin{equation}
I_g^t = \texttt{TextToImage}(P^t),
\end{equation}
where \texttt{TextToImage} indicates the text-to-image model Stable-Diffusion-V1.5~\cite{Rombach2022SD} unless otherwise stated.

\noindent\textbf{Prompt Generation.}
By comparing the differences between the current generated Image $I_g^t$ and the reference image $I_r$, candidate prompts are produced as textual gradients to update the current reverse prompt. Figure~\ref{fig:prompt-generation} presents two frameworks for prompt generation. Directly comparing two images with VLMs to generate candidate prompts exceeds their capabilities. Inspired by the chain-of-though (CoT) strategy~\cite{wei2022chain}, we propose a two-step process called vanilla prompt generation: VLMs first analyze the differences between the reference and generated images, and LLMs subsequently create candidate reverse prompts based on these differences. This strategy leverages both the image analysis capabilities of VLMs alongside the summary capabilities of LLMs. Furthermore, we observe a performance gap between open-source VLMs and GPT-4V, as open-source VLMs lacking comprehensiveness in describing differences. To mitigate this gap, we propose an enhanced prompt generation process in which VLMs first generate content and style descriptions for each image, and LLMs then compare the differences between these descriptions. Notably, when using GPT-4V, the cost of vanilla prompt generation is less than one dollar per case, while enhanced prompt generation costs approximately four dollars per case, with only minimal improvements observed. Therefore, our framework selection is driven by cost-effectiveness: we utilize the vanilla framework for GPT-4V and the enhanced framework for open-source VLMs.

At the $t$-th iteration of vanilla prompt generation, VLMs first compare the differences between the generated image $I_g^t$ and the reference image $I_r$, i.e.,
\begin{equation}
   P_{\text{diff}}^{t} = \{ P_{\text{diff}}^{t,0}, P_{\text{diff}}^{t,1}, \dots \} = \texttt{VLM}(I_g^t, I_r, \texttt{temp}),  
\end{equation}
where \texttt{temp} indicates the template or instruction used to compare image difference (See details in Appendix~\ref{appendix:prompt_generator}). An example of $P_{\text{diff}}^{t,1}$ is ``\textit{Image 1 depicts a cat with a more stylized and artistic rendering, showcasing a softer texture and a more whimsical feel with its exaggerated large blue eyes and a light blue bow tie that has a silky texture}". These descriptions comprehensively analyze the differences between two images from  content, style, color, and composition. Next, we employ LLMs to generate a candidate prompt for minimizing each difference, i.e.,  
\begin{equation}
   P_{\text{cand}}^{t} = \{P_{\text{cand}}^{t,0}, P_{\text{cand}}^{t,1}, \dots \} = \texttt{LLM}( P_{\text{diff}}^{t}, P^t, \texttt{temp}),
\end{equation}
where \texttt{temp} indicates the template or instruction used to generate candidate prompts (See details in Appendix~\ref{appendix:prompt_generator}). These generate candidate prompts are actually suggestions for minimizing each of the above-mentioned differences. An example of $P_{\text{cand}}^{t}$ is ``\{\textit{stylized artistic rendering of a cat;  exaggerated large blue eyes; light blue silky bow tie; smooth fur texture; cool tone background; serene mood; whimsical feel; illustrative and fantastical style; soft texture visual; monochromatic color scheme}\}". For the enhanced framework, it primarily enhances image difference descriptions, while the overall framework remains similar to the vanilla version. See additional  details about the enhanced framework in  Appendix~\ref{appendix:enhanced_framework}.


\noindent\textbf{Prompt Selection.}
Similar to the motivation of controlling training dynamics with learning rate and batch size in deep learning, we can not employ all candidate prompts to update the current reverse prompts because: 1) not all candidate prompts are accurate and 2) the current prompt probably can be replaced by the candidate prompts. Motivated by \cite{clip-interrogator}, we use a greedy prompt selection algorithm to identify useful candidate prompts, as shown in Algorithm~\ref{alg:clip_filter}. The similarity between a reverse prompt and the reference image, i.e., \texttt{ClipSim}, is calculated by the cosine similarity between text and image embeddings from the CLIP model~\cite{radford2021clip}. At the $t$-th iteration, the current prompt $P^t$ and the candidate prompts $P_{\text{cand}}^t$ are combined to form a new set of candidates $P_{\text{cand}}$, ensuring a fair comparison between previously selected prompts and the newly generated candidate prompts. Then, through the prompt selection process, the combination that maximizes CLIP similarity is selected as the updated reverse prompt $P^{t+1}$.

\begin{algorithm}[!t]
    \small
    \caption{The Greedy Prompt Selection Algorithm.}
    \label{alg:clip_filter}
    \begin{algorithmic}[1]
        \STATE{Current prompt $P^t$, Candidate prompts $P_{\text{cand}}^t$, Reference image $I_r$, Similarity metric \texttt{ClipSim}, Similarity score $s$.}
        \STATE{$P^{t+1} = \varnothing;~s_{\text{max}} = \texttt{ClipSim}(I_r, P^t);~P_{\text{cand}} = P^t + P_{\text{cand}}^t$}
        \WHILE{$P_{\text{cand}}^t \not = \varnothing$}
        \STATE{$p_{\text{opt}} = \mathop{\arg\max}\limits_{p \in P_{\text{cand}}}\ \texttt{ClipSim}(I_r, P^{t+1} + p)$}
        \STATE{$s = \texttt{ClipSim}(I_r, P^{t+1} + p_{\text{opt}})$}
        \IF{$s \geq s_{max}$}
        \STATE{$P^{t+1} = P^{t+1} + p_{\text{opt}};~s_{max} = s$} 
        \STATE{$P_{\text{cand}} = P_{\text{cand}} - p_{\text{opt}}$}
        \ELSE
        \STATE{\textbf{Break}}
        \ENDIF
        \ENDWHILE
        \RETURN{$P^{t+1}$}
    \end{algorithmic}
\end{algorithm}

\subsection{Novel Image Generation}
\label{sec:method:novel-image}
\begin{figure}[!t]
\centering
\begin{subfigure}{0.32\linewidth}\includegraphics[width=1\linewidth]{figure/showcase-a.pdf}
\caption{Reference Image}
\end{subfigure}
\begin{subfigure}{0.32\linewidth}\includegraphics[width=1\linewidth]{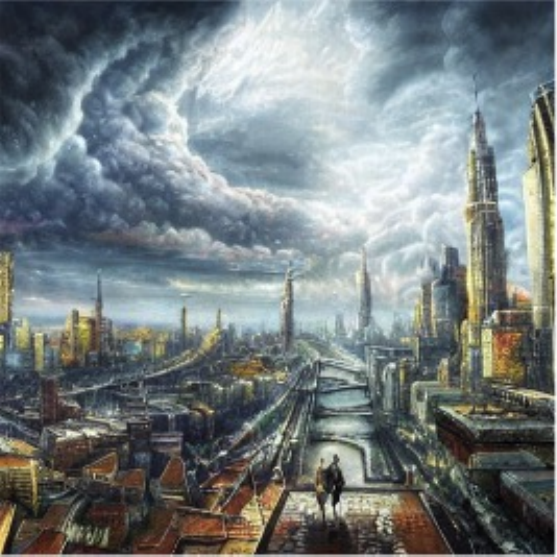}
\caption{Novel Content}
\end{subfigure}
\begin{subfigure}{0.32\linewidth}\includegraphics[width=1\linewidth]{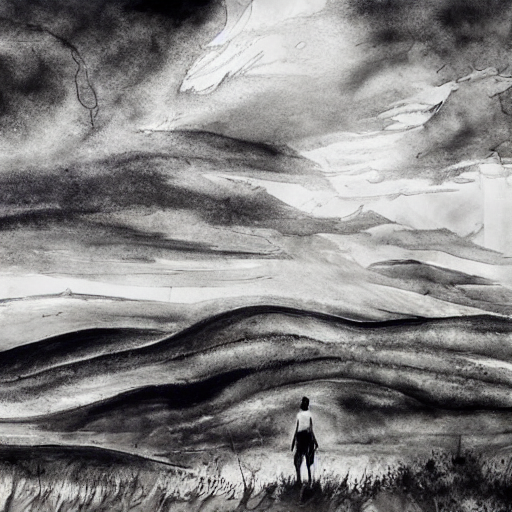}
\caption{Novel Style}
\end{subfigure}
\caption{Illustration of novel image generation. More details are provided in Appendix~\ref{appendix:novel_generation_details}.
}
\label{fig:novel-image}
\vspace{-2mm}
\end{figure}
Image reverse prompt engineering enhances our understanding of the complex relationships between textual prompts and images, facilitating research in text-to-image generation. Beyond merely recreated reference images, this technique serves practical applications by categorizing reverse prompts into content and style and helps users determine what to retain or modify to generate novel images. For the reverse prompts generated through vanilla prompt generation, we classify them using an additional LLM. In contrast, for the results generated through the enhanced prompt generation, the classification process is already integrated into the generation process. Figure~\ref{fig:novel-image} illustrates an example of generating novel images. For the novel content image, we retain all style-related reverse prompts and replace ``\textit{landscape}" in the content-related prompt with the desired ``\textit{cityscape}". Similarly, we only need to replace ``\textit{digital painting}" in the style-related reverse prompts with ``\textit{ink painting}" to generate a novel style image.

\begin{table*}[!t]
\centering 
\caption{\textbf{Results of the image and prompt fidelity comparisons}. The image fidelity metrics are assessed by running each experiment three times with three different seeds to calculate mean and variance. For our ARPO variants, ARPO (VILA, LLaMA2) indicates that we use VILA as \texttt{VLM} and LLaMA2 as \texttt{LLM} in prompt generation. Stable-Diffusion-V1.5 indicates that both reverse prompts and recreated images are generated using Stable-Diffusion-V1.5. Meanwhile, SDXL and PixArt-$\alpha$ use images generated from reverse prompts based on Stable-Diffusion-V1.5 for evaluation. 
}
\label{tab:quantitative}
\setlength\tabcolsep{2pt}
\resizebox{1\linewidth}{!}{
\begin{tabular}{lcccccccccc}
\toprule
 \multirow{2}{*}{Method} & \multirow{2}{*}{\texttt{CLIP-T}} & \multicolumn{3}{c}{Stable-Diffusion-V1.5} & \multicolumn{3}{c}{SDXL} & \multicolumn{3}{c}{PixArt-$\alpha$} \\ \cmidrule(lr){3-5} \cmidrule(lr){6-8} \cmidrule(lr){9-11}
& & \texttt{CLIP-I} & \texttt{DINO}  & \texttt{ViT} & \texttt{CLIP-I} & \texttt{DINO}  & \texttt{ViT} & \texttt{CLIP-I} & \texttt{DINO}  & \texttt{ViT}  \\ \hline
\multicolumn{11}{l}{\textbf{Gradient-based Methods}}               \\ \hline
GCG \cite{zou2023GCG, williams2024prompt} &  22.98& 59.69{\small $\pm$0.01} & 25.64{\small $\pm$0.34} & 22.22{\small $\pm$1.42} & 64.28{\small $\pm$0.28} & 33.87{\small $\pm$3.14} & 28.14{\small $\pm$2.50} & 62.10{\small $\pm$0.04} & 32.06{\small $\pm$0.01} & 26.60{\small $\pm$1.15}\\
AutoDAN \cite{zhu2023autodan, williams2024prompt} &  23.37& 59.69{\small $\pm$0.11} & 30.93{\small $\pm$0.72} & 26.39{\small $\pm$2.19} & 64.29{\small $\pm$0.10} & 33.87{\small $\pm$0.26} & 28.90{\small $\pm$0.21} & 64.07{\small $\pm$0.05} & 34.03{\small $\pm$0.21} & 29.25{\small $\pm$0.22}\\
Random Search \cite{zou2023GCG, williams2024prompt} &  22.59& 60.72{\small $\pm$0.23} & 27.30{\small $\pm$0.05} & 21.00{\small $\pm$0.30} &62.82{\small $\pm$0.09} & 33.76{\small $\pm$0.15} & 26.14{\small $\pm$0.04} & 61.52{\small $\pm$0.66} & 30.83{\small $\pm$1.49} & 23.15{\small $\pm$0.18} \\
PEZ~\cite{wen2024hard} & 21.81 & 72.04{\small $\pm$1.97} & 40.65{\small $\pm$0.50} & 39.47{\small $\pm$1.00} & 68.37{\small $\pm$0.32} & 37.73{\small $\pm$0.24} & 33.28{\small $\pm$0.01} & 67.78{\small $\pm$0.05} & 35.89{\small $\pm$0.01} & 32.73{\small $\pm$0.19} \\
PH2P~\cite{mahajan2024prompting} & 23.76 & 76.27{\small $\pm$0.39} & 46.27{\small $\pm$0.99} & 44.28{\small $\pm$0.46} & 72.38{\small $\pm$0.64} & 44.16{\small $\pm$0.22} & 38.93{\small $\pm$0.06} & 69.21{\small $\pm$0.73} & 41.01{\small $\pm$0.04} & 35.77{\small $\pm$0.17} \\ \hline
\multicolumn{11}{l}{\textbf{Image Captioning Methods}} \\ \hline
BLIP2~\cite{li2023blip2} & 26.42  & 75.81{\small $\pm$0.22}  & 46.06{\small $\pm$0.24} & 44.74{\small $\pm$0.16} & 75.72{\small $\pm$0.02} & 49.60{\small $\pm$0.25} & 44.37{\small $\pm$0.11} & 74.69{\small $\pm$0.03} & 46.60{\small $\pm$0.02} & 43.41{\small $\pm$0.33} \\
ShareCaptioner~\cite{chen2023sharegpt4v} & 26.72  & 76.28{\small $\pm$0.05}  & 48.11{\small $\pm$0.27} & 44.97{\small $\pm$0.71} & 77.47{\small $\pm$0.21} & 51.70{\small $\pm$0.19} & 47.05{\small $\pm$0.04} & 76.84{\small $\pm$0.02} & 50.61{\small $\pm$0.01} & 47.57{\small $\pm$0.01} \\
LLaVA-Next~\cite{liu2024llava-next} & 26.54 & 76.13{\small $\pm$0.03} & 45.90{\small $\pm$0.03} & 44.28{\small $\pm$0.46} & 75.53{\small $\pm$0.06} & 51.28{\small $\pm$0.56} & 46.94{\small $\pm$0.02} & 75.97{\small $\pm$0.02} & 51.57{\small $\pm$0.02} & 46.94{\small $\pm$0.17}\\
VILA~\cite{lin2023vila} & 27.76  & 76.27{\small $\pm$0.26}  & 48.53{\small $\pm$0.22} & 45.83{\small $\pm$0.05} & 76.61{\small $\pm$0.16} & 52.95{\small $\pm$0.12} & 48.83{\small $\pm$0.13} & 77.21{\small $\pm$0.05} & 51.32{\small $\pm$0.16} & 46.94{\small $\pm$0.13} \\
GPT-4V~\cite{achiam2023gpt} & 28.39  & 78.14{\small $\pm$0.07}  & 50.40{\small $\pm$0.18} & 45.86{\small $\pm$0.14} & 77.81{\small $\pm$0.06} & 52.55{\small $\pm$0.04} & 48.19{\small $\pm$0.21} & 77.75{\small $\pm$0.12} & 52.91{\small $\pm$0.42} & 48.29{\small $\pm$0.72} \\ \hline
\multicolumn{11}{l}{\textbf{Data-driven Methods}}                              \\ \hline
CLIP-Interrogator~\citep{clip-interrogator} & 30.56  & 80.62{\small $\pm$0.13}  & 50.43{\small $\pm$0.14} & 46.30{\small $\pm$0.34} & 79.18{\small $\pm$0.80} & 52.61{\small $\pm$0.46} & 48.51{\small $\pm$0.42} & 77.88{\small $\pm$0.01} & 49.23{\small $\pm$0.04} & 45.02{\small $\pm$0.02} \\ \hline
\multicolumn{11}{l}{\textbf{ARPO (Ours)}}                \\ \hline
 ARPO (VILA, Mistral)                 & 35.03  & 81.55{\small $\pm$0.04}  & 52.37{\small $\pm$0.51} & 49.22{\small $\pm$0.43} & 80.89{\small $\pm$0.01} & 54.55{\small $\pm$0.02} & 50.03{\small $\pm$0.04} & 78.69{\small $\pm$0.01}& 51.71{\small $\pm$0.08} & 48.08{\small $\pm$0.48}  \\
ARPO (VILA, LLaMA2) & 35.14  & 81.38{\small $\pm$0.01}  & 53.37{\small $\pm$0.24} & 49.35{\small $\pm$0.14} & 80.31{\small $\pm$0.02} & 54.67{\small $\pm$0.07} & 50.65{\small $\pm$0.17} & 78.95{\small $\pm$0.02} & 52.03{\small $\pm$0.01} & 49.40{\small $\pm$0.19}\\
ARPO (LLaVA-Next, Mistral) & 35.27  & 81.50{\small $\pm$0.12}  & 53.69{\small $\pm$0.23} & 50.26{\small $\pm$0.14} & 80.90{\small $\pm$0.02} & 54.42{\small $\pm$0.07} & 51.03{\small $\pm$0.16} & 78.78{\small $\pm$0.01} & 52.93{\small $\pm$0.19} & 49.51{\small $\pm$0.06} \\
ARPO (LLaVA-Next, LLaMA2) & \textbf{35.58}  & \textbf{83.01{\small $\pm$0.02}}  & \textbf{54.00{\small $\pm$0.45}} & \textbf{51.40{\small $\pm$0.27}} & \textbf{81.57{\small $\pm$0.09}} & \textbf{55.24{\small $\pm$0.01}} & \textbf{52.57{\small $\pm$0.12}} & \textbf{80.40{\small $\pm$0.07}} & \textbf{53.37{\small $\pm$0.12}} & \textbf{50.43{\small $\pm$0.07}} \\
\bottomrule
\end{tabular}}
\end{table*}

\section{Experiments}

\subsection{Dataset and Metrics}
Following the dataset scale of \citet{mahajan2024prompting} and \citet{williams2024prompt}, we constructed a benchmark dataset with 50 human-created images (artistic paintings from WikiArt~\citep{saleh2015wikiart} and photographs from~\cite{jing2020dynamic}) and 50 AI-generated images from DiffusionDB~\citep{wang2022diffusiondb}. This dataset ensures a diverse set of test images with different styles.
We employ two types of metrics: 1) \textbf{prompt fidelity} measures the cosine similarity between the CLIP \citep{radford2021clip} embeddings of the reverse prompt and the reference image, denoted as \texttt{CLIP-T}; 2) \textbf{image fidelity} measures the cosine similarity between the embeddings of the reference image and the recreated image using the reverse prompt. Specifically, we consider three embedding extractors, CLIP~\citep{radford2021clip}, DINO~\citep{caron2021DINO}, and ViT~\cite{dosovitskiy2020vit}, and the metrics are referred to as \texttt{CLIP-I}, \texttt{DINO}, and \texttt{ViT}, respectively.
Considering the inherent randomness in text-to-image models, all image fidelity metrics are assessed in each experiments three times using different random seeds to calculate the mean and variance. 

\subsection{Quantitative Comparisons}

We compare the following methods for image reverse prompt engineering: 1) \textbf{gradient-based methods} (PEZ~\cite{wen2024hard}, GCG~\cite{zou2023GCG,williams2024prompt}, AutoDAN~\cite{zhu2023autodan,williams2024prompt}, Random Search~\cite{andriushchenko2023randomsearch, williams2024prompt}, PH2P~\cite{mahajan2024prompting}) 2) \textbf{image captioning methods} (BLIP2~\cite{li2023blip2}, ShareCaptioner~\cite{chen2023sharegpt4v}, LLaVA-Next~\cite{liu2024llava-next}, VILA~\cite{lin2023vila}, GPT-4V~\cite{achiam2023gpt}); 3) \textbf{data-driven methods} (CLIP-Interrogator~\citep{clip-interrogator}); and 4) \textbf{our ARPO variants}. 
As shown in Table~\ref{tab:quantitative}, GCG, AutoDAN, Random Search, PEZ and PH2P achieved poor results, indicating that gradient-based methods lack accuracy in the IRPE task. This may be due to the loss of useful information during the discretization process, which results in an inability to comprehensively describe the reference image. Moreover, advanced VLMs typically excel in generating better reverse prompts. However, attempting to directly utilize general VLMs for this purpose does not yield competitive results, even with the considerable capabilities of GPT-4V. This limitation may arise from their preference for lengthy and complex descriptions, which pose challenges for adaptation by text encoders in Text-to-Image (T2I) models. It is noteworthy that the CLIP-Interrogator outperforms GPT-4V, primarily due to the utilization of a high-quality dataset tailored for specific purposes. Nevertheless, the limited scope of such a specific dataset hinders achieving significantly better results, as reverse prompts generated by CLIP-Interrogator are constrained to the existing entries in the dataset. Compared to baseline methods, the proposed ARPO consistently achieves significantly better results across various choices of VLM/LLM components. This highlights the effectiveness and robustness of our method for image reverse prompt engineering. We also conduct experiments on human hand-crafted prompts, commerical services and closed-source (GPT4-V and GPT4) ARPO method on a subset of 40 images due to the high huamn labor and cost. Details about the experiments are provide in the Appendix~\ref{appendix:subset}.


To showcase the adaptability and generalization capability of our ARPO method to different T2I models, we explore two additional popular models, SDXL~\cite{podell2023sdxl} and PixArt-$\alpha$~\cite{chen2023pixart}, for comparison with baseline methods. Using the prompts generated for Stable-Diffusion-v1.5 in previous experiments, we input them into SDXL and PixArt-$\alpha$ to generate new images and assess fidelity. As shown in Table~\ref{tab:quantitative}, the VLMs method and CLIP-Interrogator, being independent of T2I models, exhibit excellent adaptability and generalization. In contrast, PEZ and PH2P, which rely on T2I models for prompt tuning, are highly model-dependent. While ARPO also interacts with T2I models during generation, experimental results confirm that its outputs transfer well across models while maintaining state-of-the-art performance. 
We further analyze the impact of the T2I models choice for ARPO and image recreation. As shown in Table~\ref{tab:different_models}, using the same model for reverse prompt and image creation improves fidelity, suggesting that ARPO can learn prompts specific to T2I models. Additionally, the reverse prompts exhibit transferability across different models, highlighting the potential of small-scale models to generate effective reverse prompts for larger models.

\begin{table}[t]
\centering
\caption{The performance of APRO using different T2I models.
}
\label{tab:different_models}
\resizebox{0.90\linewidth}{!}{
\begin{tabular}{lcccc}
\toprule
Generation & ARPO & \texttt{CLIP-I} & \texttt{DINO}  & \texttt{ViT} \\ \hline
\multirow{3}{*}{SD-V1.5} & SD-V1.5 & \textbf{83.01{\small $\pm$0.02}} & \textbf{54.00{\small $\pm$0.45}} & \textbf{51.40{\small $\pm$0.27}} \\
& SDXL & 82.20{\small $\pm$0.03} & 53.86{\small $\pm$0.51} & 51.03{\small $\pm$0.15} \\
& PixArt-$\alpha$ & 81.74{\small $\pm$0.12} & 53.16{\small $\pm$0.06} & 50.53{\small $\pm$0.05} \\ \hline
\multirow{3}{*}{SDXL} & SD-V1.5 & 81.57{\small $\pm$0.09} & 55.24{\small $\pm$0.01} & 52.57{\small $\pm$0.12} \\
& SDXL & \textbf{82.36{\small $\pm$0.04}} & \textbf{55.98{\small $\pm$0.01}} & \textbf{53.07{\small $\pm$0.12}} \\ 
& PixArt-$\alpha$ & 80.79{\small $\pm$0.05} & 54.35{\small $\pm$0.58} & 51.12{\small $\pm$0.42}\\ \hline
\multirow{3}{*}{PixArt-$\alpha$} & SD-V1.5 & 80.40{\small $\pm$0.07} & 53.37{\small $\pm$0.12} & 50.43{\small $\pm$0.07} \\
& SDXL & 79.99{\small $\pm$0.12} & 53.81{\small $\pm$0.01} & 50.05{\small $\pm$0.33}\\
& PixArt-$\alpha$ & \textbf{81.02{\small $\pm$0.18}} & \textbf{53.99{\small $\pm$0.14}} & \textbf{51.06{\small $\pm$0.22}}\\
\bottomrule
\end{tabular}
}
\end{table}

\subsection{User Study}
\begin{figure}[t]
    \centering
    \includegraphics[width=0.75\linewidth]{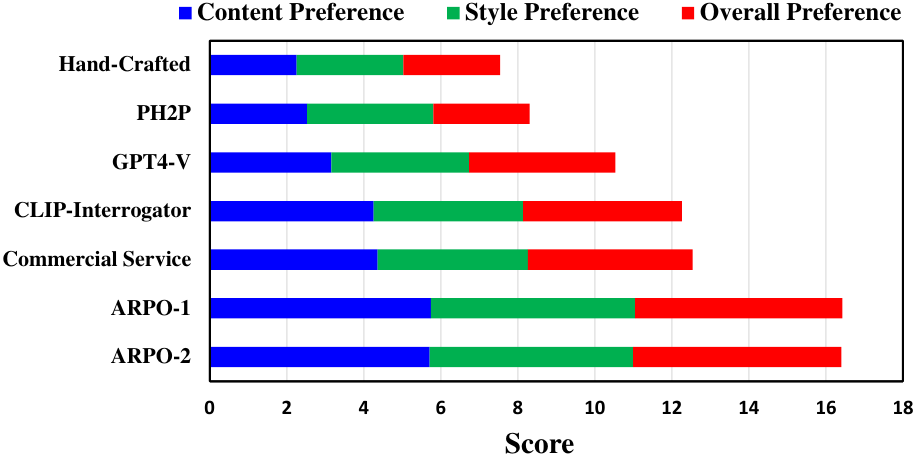}
    \caption{\textbf{The results of our user study.} ARPO-1 uses the combination of LLaVA-Next and LLaMA2, while ARPO-2 uses the combination of GPT-4V and GPT-4 for prompt generation.}
    \label{fig:user_study}
\end{figure}
We conduct a user study to compare images generated by six methods: hand-crafted, PH2P, GPT-4V, CLIP-Interrogator, a commercial service, and two ARPO variants. The study involved 40 cases, with 50 users ranking the methods based on similarity to the reference image, considering content, style, and overall preference. The first rank received a score of 7, and the seventh rank received a score of 1. As shown in Figure~\ref{fig:user_study}, both ARPO variants achieved comparable performances, clearly surpassing the hand-crafted method, PH2P, GPT-4V, CLIP-Interrogator, and the commercial service.

\begin{table}[t]
\centering 
\caption{\textbf{Ablation studies} on prompt generation and  selection.}
\label{tab:ablation}
\resizebox{\linewidth}{!}{
\begin{tabular}{lcccc}
\toprule
Method &\texttt{CLIP-T} & \texttt{CLIP-I} & \texttt{DINO}  & \texttt{ViT}    \\ \hline
ARPO & 35.58  & 83.01{\small $\pm$0.02}  & 54.00{\small $\pm$0.45} & 51.40{\small $\pm$0.27} \\ \hline
\quad VLMs Only & 20.17  & 67.02{\small $\pm$0.35} & 36.07{\small $\pm$0.57} & 32.07{\small $\pm$0.42} \\
\quad w/o Enhanced & 33.04  & 78.54{\small $\pm$0.18}  & 50.96{\small $\pm$0.25}  & 47.98{\small$\pm$0.11} \\
\quad w/o Selection & 27.09  & 78.65{\small $\pm$0.26}  & 49.76{\small $\pm$0.05} & 47.82{\small $\pm$0.16} \\
\quad w/o Combination & 30.40  & 79.74{\small $\pm$0.08}  & 50.67{\small $\pm$0.26} & 48.94{\small $\pm$0.35} \\ 
\bottomrule
\end{tabular}}
\end{table}

\begin{figure}[t]
\centering
\begin{subfigure}{0.4\linewidth}\includegraphics[width=1\linewidth]{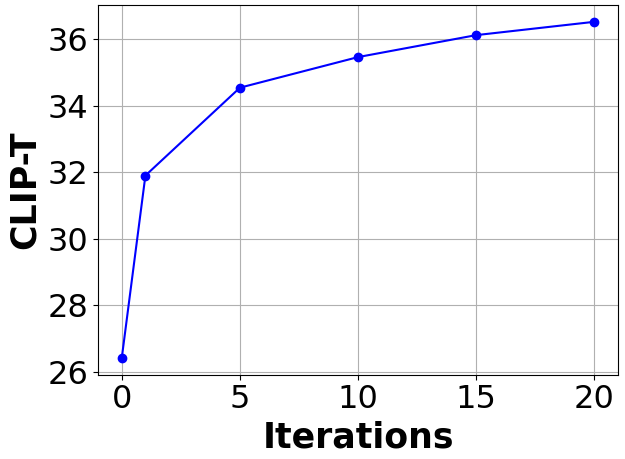}
\caption{\texttt{CLIP-T}}
\end{subfigure}
\begin{subfigure}{0.4\linewidth}\includegraphics[width=1\linewidth]{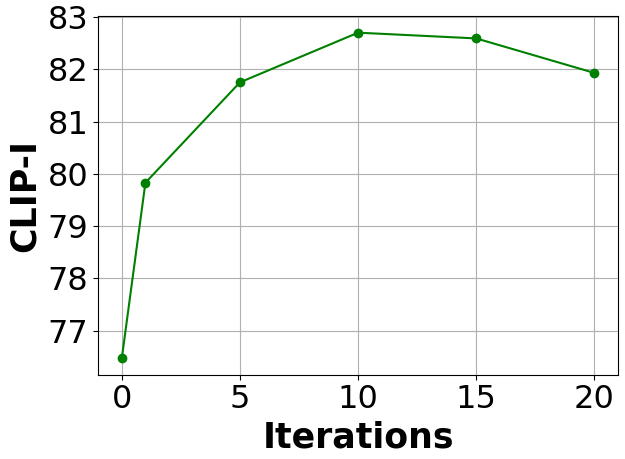}
\caption{\texttt{CLIP-I}}
\end{subfigure}
\vfill
\begin{subfigure}{0.4\linewidth}\includegraphics[width=1\linewidth]{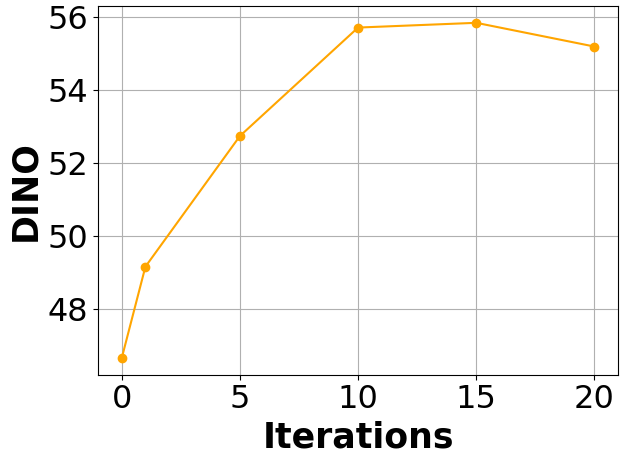}
\caption{\texttt{DINO}}
\end{subfigure}
\begin{subfigure}{0.4\linewidth}\includegraphics[width=1\linewidth]{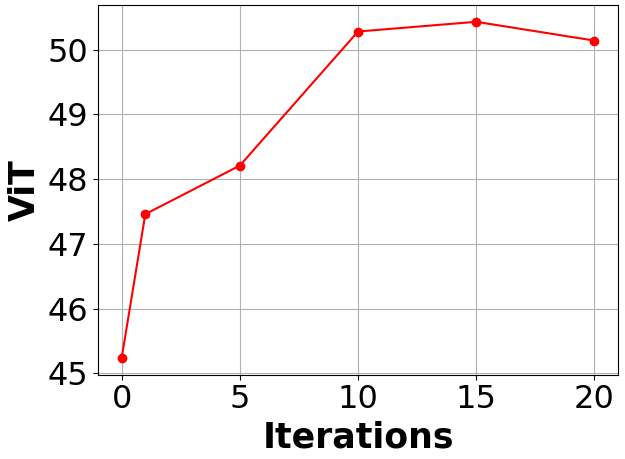}
\caption{\texttt{ViT}}
\end{subfigure}
\caption{The \textbf{influence} of \textbf{optimization steps}.}
\label{fig:steps}
\end{figure}

\subsection{Ablation Studies}

\paragraph{Prompt Generation}
To generate candidate prompts, we propose two prompt generation frameworks to elicit the ability of LLMs/VLMs. VLMs Only indicates that directly comparing images and generating candidate prompts with single VLM. The poor results shown in Table~\ref{tab:ablation} demonstrate that this approach exceeds the capabilities of VLMs and highlight the importance of our two-step compare-and-generating strategy to utilize the image analysis ability of VLMs and the summary ability of LLMs. Furthermore, we notice the performance gap between open-source VLMs and GPT-4V. As shown in Table~\ref{tab:ablation}, the describe-then-compare strategy in the enhance framework enables VLMs to provide more comprehensive details, improving both text and image fidelity, and significantly mitigating the performance degradation caused by some weaker VLMs.

\noindent\textbf{Prompt Selection }
As shown in Table~\ref{tab:ablation}, using all generated candidate prompts to update the reverse prompt leads to significant performance degradation. This result indicate that not all candidate prompts are correct or helpful to reduce differences between reference and generated images, which highlight the importance of our selection method. In our prompt selection algorithm, an important strategy is to combine the current reverse prompt with the candidate prompts and then fairly select them to generate a update reverse prompt. As the results shown in Table~\ref{tab:ablation}, this strategy greatly improves the performance of our method. This indicates that during the optimization process, the prompt generation process can produce better candidate prompts to replace previously generated results, efficiently improving the accuracy of the reverse prompt.

\noindent\textbf{Optimization Steps }
In Figure~\ref{fig:steps}, we illustrate performances across different numbers of optimization steps. Specifically, a maximum of 10 steps is typically sufficient in most cases.  When running more iterations, there may be some performance fluctuations, but differences in the generated images are typically imperceptible to the human eye until many more optimization steps are taken. Therefore, employing an early stop strategy would be useful, otherwise it is also easy for us to empirically determine the proper number of steps in practice.  Notably, even with two optimization steps, it already outperforms all other methods.

\subsection{Qualitative Comparisons}
In this section, we qualitatively compare our ARPO method with several baseline methods across three application scenarios: image recreation, image modification, and image fusion. The baseline methods include the hand-crafted method, PH2P using discrete prompt tuning, image caption of a open-source VLMs LLaVA-Next and a closed-source VLMs GPT4-V, and the CLIP-Interrogator method based on prior dataset. 

\begin{figure}[t]
    \centering
    \includegraphics[width=1\linewidth]{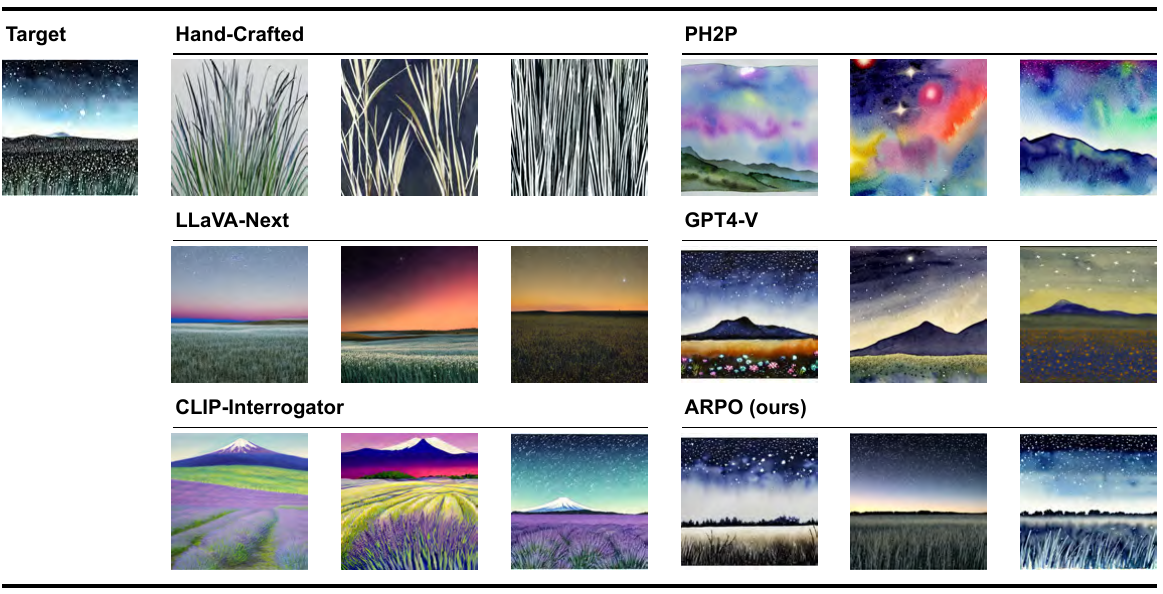}
    \caption{Illustration of recreated images using the reverse prompts from different methods.}
    \label{fig:compare}
\end{figure}

\noindent\textbf{Image Recreation}
Image recreation involves generating an image similar to the reference image using a reverse prompt. In this experiment, we generate three recreated images with different random seeds using the reverse prompt from different methods. The random seeds for different methods are the same. As shown in the Figure~\ref{fig:compare}, compared to these baseline methods, our ARPO method better preserves various elements of the reference image, such as more accurately retaining the image content and more closely reproducing the image style and color. The image fidelity results corresponding to the recreated images and more experimental results are provided in Appendix~\ref{appendix:visualization_and_fidelity} and \ref{appendix:image_reconstruction}.

\begin{figure}[!t]
    \centering
    \includegraphics[width=1\linewidth]{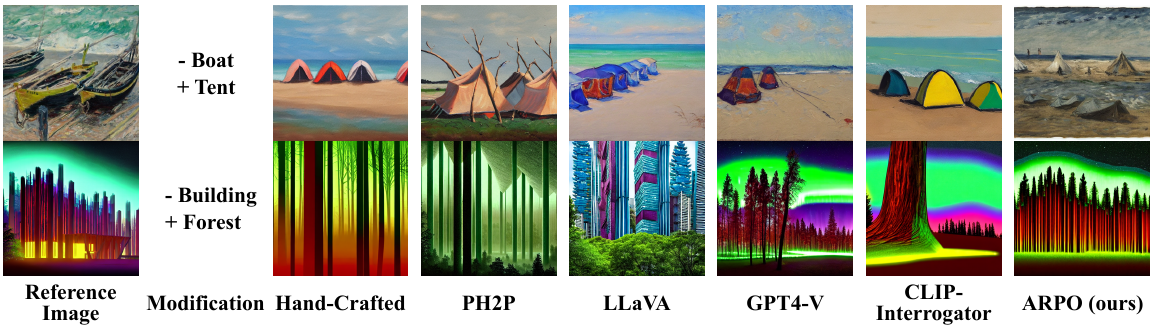}
    \caption{Comparison of editing reverse prompts for novel creations. For modification, "- Boat + Tent" means we replace all instances of the word "boat" in reverse prompt with the word "tent".}
    \label{fig:control_compare}
\end{figure}

\begin{figure}[!t]
    \centering
    \includegraphics[width=1\linewidth]{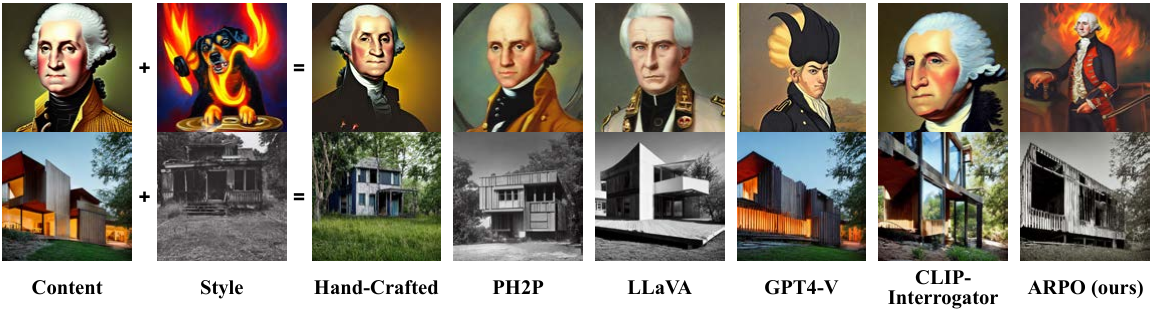}
    \caption{Comparison of image fusion. A new image is generated by merging style-related prompts from the "Style" image with content-related prompts from the "Content" image.}
    \label{fig:image_fusion}
\end{figure}

\noindent\textbf{Image Modification \& Fusion} We compare the ability of generating novel images using reverse prompts from our ARPO method against baseline methods. We conduct experiments in two scenario: image modification and image fusion. Image modification involves editing part of the reverse prompt to generate novel images, while image fusion addresses a more challenging application scenario by extracting style-related information from one image and content-related information from another, and then combining them to create a new image. 

We first analyze the experience of editing prompts. Due to the lack of readability in the prompts generated by gradient-based methods, it is challenging to extract the necessary information for editing. Therefore, we follow the prompt strategy in PEZ, using the prompt format of `` a \{\} in the style of \{\}" for image modification and concatenating two reverse prompt for image fusion. For image captioning method, their reverse prompts often mix style and content into a single sentence and contain repetitive and useless information, requiring a significant amount of time to extract and edit the relevant information. For hand-crafted prompts, CLIP-Interrogator, and our ARPO method, since the prompts are provided in the form of tags, we can simply classify them using LLMs and then merge them, making the operation more convenient.

The image comparison results for these two tasks are shown in Figure~\ref{fig:control_compare} and \ref{fig:image_fusion}. More results are provided in Appendix~\ref{appendix:comparison_image_modification} and \ref{appendix:image_fusion}. We can draw the following observation: 1) Our method allows the novel image to more accurately maintain the desired elements of the reference image while precisely modifying the intended elements; 2) our method ensures that style and content are separated without interfering with each other.

\section{Conclusion and Future Work}
In this paper, we addressed the challenge of image reverse prompt engineering and introduced an automatic reverse prompt optimization method. Using the obtained reverse prompts, we can easily generate novel images with content or style similar to the reference image. 
Our method consistently outperforms many baseline approaches, including hand-crafted methods, image caption methods, and data-driven methods, as demonstrated by both quantitative and qualitative results. This research underscores the promising potential of reverse prompt engineering for text-to-image. 
\textbf{Limitation and Future Work.} Our ARPO method involves large model inference and multi-step optimization, leading to higher computational costs than baseline methods. Future work could focus on improving its efficiency.



{
    \small
    \bibliographystyle{ieeenat_fullname}
    \bibliography{main}
}

\clearpage
\onecolumn
\appendix

\section{Computational Resource}
For the ARPO method with closed-source models and vanilla prompt generation framework, we utilize GPT4-V (gpt-4-vision-preview) and GPT4 (gpt-4-1106-preview) as the VLM and LLM. For the ARPO with open-source models and enhanced prompt generation process, we try to utilize VILA (VILA-13b, CC-BY-NC-4.0) or LLaVA-Next (llava-v1.6-vicuna-13b, LLAMA 2 Community License) as our VLM and LLaMA2 (Llama-2-13b-chat-hf, Llama 2 Community License) or Mistral (Mistral-7B-Instruct-v0.2, APACHE-2.0) as our LLM. We employ Stable-Diffusion-V1.5 (stable-diffusion-v1-5, creativeml-openrail-m), SDXL (stable-diffusion-xl-base-1.0, openrail++), and PixArt-$\alpha$ (PixArt-XL-2-512$\times$512,openrail++) as our Text-to-Image model to test our method can handle different image generators. In the prompt initialization step, we use BLIP2 (blip2-opt-2.7b, MIT License) to get a concise image caption. In the step of prompt selection, we use CLIP-ViT-L-14 (MIT License) to calculate the cosine similarity between text and image CLIP embeddings. 

We deploy our closed-source models version on a single NVIDIA 4090 24G GPU and our open-source models version on two NVIDIA A100 40G GPUs. The closed-source version takes about 20 seconds and costs approximately 0.07 USD per iteration. In contrast, due to the additional image description process in the enhanced prompt generation framework, the open-source version takes around 90 seconds per iteration but is completely free.

\section{More details about Prompt Generation}
\subsection{Vanilla Framework}
\label{appendix:prompt_generator}
In the framework of the vanilla prompt generation process, we first utilize the VLM model to compare the difference between the generated image and the reference image. The template used by VLM model to compare image differences is shown in Figure~\ref{fig:image_difference}. An example of difference description generated by GPT4-V is shown in Figure~\ref{fig:difference_example}. Then we employ LLM to generate candidate prompts for minimizing each of the identified differences. The template used by LLM is shown in Figure~\ref{fig:prompt_candidate} and a candidate prompts example generated by GPT-4 is shown in Figure~\ref{fig:candidate_example}.

\begin{figure}[H]
    \begin{tcolorbox}[title=Template used by the VLLM to compare image difference.]
        The first image is Image 1 and the second image is Image 2. You need to describe the difference between Image 1 and Image 2. Let's think step by step.
    \end{tcolorbox}
    \caption{Template used by the VLLM to compare image difference between the reference image and the generated image.}
    \label{fig:image_difference}
\end{figure}

\begin{figure}[H]
    \begin{tcolorbox}[title=An example of difference description]
        Image 1 and Image 2 both feature a cat wearing a bow tie, but there are notable differences. Image 1 depicts a cat with a more stylized and artistic rendering, showcasing a softer texture and a more whimsical feel with its exaggerated large blue eyes and a light blue bow tie that has a silky texture. The cat's fur appears smooth and the overall image has a cool tone, contributing to a serene mood. In contrast, Image 2 presents a cat with a realistic photograph quality, with sharper textures seen in the fur, a solid black and white color pattern on the fur, and more natural but still vivid blue eyes. The bow tie here is darker and less reflective, suggesting a different material. The background of Image 2 is a neutral brown, which contrasts with the cooler, more monochromatic background of Image 1. The mood in Image 2 feels more straightforward and less fanciful than in Image 1. The layout and scale of the subjects are similar, with both cats positioned centrally and occupying a comparable amount of space within the frame, but the style of Image 1 is more illustrative and fantastical, while Image 2 is more realistic and lifelike.
    \end{tcolorbox}
    \caption{An example of difference description generated by GPT4-V}
    \label{fig:difference_example}
\end{figure}

\begin{figure}[H]
    \begin{tcolorbox}[title=Template used by the LLM to generate prompt candidates.]
        Generate image promts that incorporate the following difference between Image 1 and Image 2: \{difference\}.\\
        For the specific contrasts identified in the differences between Image 1 and Image 2, the image prompts should guide the creation of images that align more closely with Image 1.\\
        The prompts should be structured as a series of keywords or short phrases, separated by commas. Please list all possible prompts in a python list format. Your answer should only contain a python list. Let's think step by step.
    \end{tcolorbox}
    \caption{Template used by the LLM to generate prompt candidates based on the difference between two image.}
    \label{fig:prompt_candidate}
\end{figure}

\begin{figure}[H]
    \begin{tcolorbox}[title=An example of candidate prompts]
        [stylized artistic rendering of a cat, exaggerated large blue eyes, light blue silky bow tie, smooth fur texture, cool tone background, serene mood, whimsical feel, illustrative and fantastical style, soft texture visual, monochromatic color scheme]
    \end{tcolorbox}
    \caption{An example of candidate prompts generated by GPT4}
    \label{fig:candidate_example}
\end{figure}

\subsection{Enhanced Framework}
\label{appendix:enhanced_framework}
Using open-source models like VILA and LLaVA-Next with relatively lower performing compared to GPT4-V to generate image differences has two significant challenges: 1) Some VLMs are trained solely on single image-text pairs and lack the capability for multi-image inference. 2) Their comparative results are often not comprehensive, capturing only partial differences. To address this shortcoming, we propose an enhanced prompt generation framework for open-source models. The enhanced framework is primarily used to improve the descriptions of image differences, while the overall framework remains similar to the vanilla version.

In this framework, we propose only using VLM to generate description for one image at a time. In order to provide a comprehensive description of the image, we divide the description of the image into two parts: content and style. Content focuses on capturing information such as the objects, characters, events, backgrounds, and actions, etc. depicted the image, while style concentrates on aspects such as artistic style, colour, media, and lighting, etc. The description process is
\begin{equation}
    D = \texttt{LLM}(I, \texttt{template})
\end{equation}
where \texttt{template} indicates the template or instruction used by the VLM to descibe image. VLM uses different templates to generate descriptions for content and style. The templates for content descriptions and style descriptions are shown in Figure~\ref{fig:content_description} and Figure~\ref{fig:style_description}. An example of content descriptions and style descriptions generated by VILA is shown in Figure~\ref{fig:description_example}. 

The VLM will generate separate descriptions for content and style for both the reference image and the generated image respectively. Following this, a LLM uses these descriptions to generate difference descriptions. It compares the content description of the reference image $D_r$ with that of the generated image $D_g^t$, and similarly for the style descriptions,
\begin{equation}
    P_{\text{diff}}^t = \texttt{LLM}(D^t_g, D_r, \texttt{template})
\end{equation}
where \texttt{template} indicates the template or instruction used by LLM to generate image differences. The details of this template is shown in Figure~\ref{fig:description_difference}. An example of content differences generated by LLaMA2 is shown in Figure~\ref{fig:difference_enhance_example}. Subsequent steps are the same as for the vanilla framework, utilizing LLM to generate candidate prompts.

\begin{figure}[H]
    \begin{tcolorbox}[title=Template used by the VLM to generate content description.]
        You are an expert in describing image, please describe the content of the image. This includes indentifying objects, environments, events, background, actions, etc. in the image.
    \end{tcolorbox}
    \caption{Template used by the VLM to generate content description.}
    \label{fig:content_description}
\end{figure}

\begin{figure}[H]
    \begin{tcolorbox}[title=Template used by the VLM to generate style description.]
        You are an expert in image analysis, please describe the style of the image. This includes identifying the medium of the image, the art style, the artist's style, the creative technique, the lighting, the colours and the resolution, etc.
    \end{tcolorbox}
    \caption{Template used by the VLM to generate style description.}
    \label{fig:style_description}
\end{figure}

\begin{figure}[H]
    \begin{tcolorbox}[title=An example of content and style descriptions.]
        \textbf{Content Descriptions} \\
        The image portrays a black cat, adorned with a blue bow tie, standing against a gray background. The cat's eyes are a striking shade of blue, adding a touch of whimsy to the scene. The bow tie, a symbol of elegance and formality, contrasts beautifully with the cat's black fur. The gray background serves to highlight the cat and its accessory, making them the focal point of the image. The cat's position and the direction of its gaze suggest it is looking directly at the viewer, creating a sense of connection between the viewer and the subject of the image. The overall composition of the image is balanced and harmonious, with each element - the cat, the bow tie, and the background - contributing to a visually pleasing scene.\\
        \hdashrule[0.5ex]{\linewidth}{1pt}{3mm 2pt}
        \textbf{Style Descriptions}\\
        The image portrays a black cat with striking blue eyes, adorned with a blue bow tie. The cat is positioned in the center of the image, facing the viewer directly. The background is a uniform gray, which serves to highlight the cat and its vibrant attire. The medium of the image is digital art, as evidenced by the crisp lines and smooth gradients. The art style is realistic, with attention to detail in the cat's features and the intricate design of the bow tie. The creative technique used is digital painting, which allows for precise control over the color palette and the overall composition of the image. The lighting is even, with no shadows or highlights, further emphasizing the cat as the focal point of the image. The colors are vivid and saturated, particularly the blue of the cat's eyes and the bow tie, which stand out against the gray background. The resolution of the image is high, allowing for a clear and detailed view of the cat and its surroundings. Overall, the image is a beautiful representation of a black cat, captured in a striking and unique pose
    \end{tcolorbox}
    \caption{An example of content and style descriptions generated by VILA.}
    \label{fig:description_example}
\end{figure}

\begin{figure}[H]
    \begin{tcolorbox}[title=Template used by the LLM to generate image differences.]
        I have descriptions of Image 1 and Image 2.\\
        The descriptions of Image 1: \{image1\}\\
        The descriptions of Image 2: \{image2\}\\
        Please identify the differences between Image 1 and Image 2 based on their descriptions. Let's think step by step.
    \end{tcolorbox}
    \caption{Template used by the LLM to generate image differences based on the image description.}
    \label{fig:description_difference}
\end{figure}

\begin{figure}[H]
    \begin{tcolorbox}[title=An example of content difference.]
        1. Color scheme: Image 1 features a predominantly gray and blue color scheme, while Image 2 has a more muted color scheme with a focus on black, white, and green.\\
        2. Subject: Image 1 features a black cat with a blue bow tie, while Image 2 features a black and white cat with striking green eyes.\\
        3. Accessories: Image 1 features a blue bow tie, while Image 2 does not have any discernible accessories.\\
        4. Background: Image 1 has a gray background, while Image 2 has a plain gray background.\\
        5. Composition: Image 1 has a more balanced composition, with the cat, bow tie, and background all contributing to a visually pleasing scene. Image 2 has a more centralized composition, with the cat positioned in the center of the frame and the background fading into the distance.\\
        6. Focal point: Image 1 features the blue bow tie as a focal point, while Image 2 features the cat's green eyes as the focal point.\\
        7. Mood: Image 1 has a more playful and whimsical mood, while Image 2 has a more curious and inquisitive mood.\\
        8. Context: Image 1 provides more context, with the background and accessories giving a sense of setting and purpose. Image 2 does not provide any additional context or information about the cat's surroundings or actions.
    \end{tcolorbox}
    \caption{An example of content difference generated by LLaMA2.}
    \label{fig:difference_enhance_example}
\end{figure}

\subsection{Trade-offs between Vanilla and Enhanced Framework}
Compared to the vanilla framework, the enhanced framework can more effectively harness the potential of VLMs, yielding better candidate prompts. However, we apply the enhanced framework exclusively to open-source models, while continuing to use the vanilla framework for GPT4-V and GPT-4. This decision is based on some disadvantages associated with the enhanced framework. Specifically, the enhanced framework not only adds an image description process but also divides the tasks into content and style categories. This design significantly increases the number of tokens input and output to VLMs and LLMs. For the OPENAI API, which charges based on the number of input and output tokens, using the enhanced framework increases the cost to 4 USD, approximately four times the expense of using the vanilla framework. Additionally, the inference time is substantially extended, requiring more than four times the duration compared to the vanilla framework.

Considering the excellent performance of GPT4-V, the benefits of using the enhanced framework are not as significant as with open-source models. Thus, we balanced performance against costs. We introduce two ARPO version: a closed-source version based on closed-source models using the vanilla framework, which offers faster processing speed but incurs API costs, and a open-source version using the enhanced framework with open-source models, which requires more processing time but is completely free. The performance of these two versions is similar, allowing users to trade-off time and cost to choose the suitable method for their needs.

\clearpage
\section{Experiments}
\subsection{Experiments of Hand-crafted, Commercial Services and Closed-source ARPO}
\label{appendix:subset}
We conduct quantitative experiments of human hand-crafted prompts and commercial services ( ImageToPromptAI and Phot.AI) and closed-source (GPT4-V and GPT4) ARPO method. Considering the extensive manpower required of human hand-crafted prompts and the expensive cost of the commercial services and OPENAI API, we randomly select a subset of 40 images for experiments. For a fair comparison, we recalculate the results for the two best performing baseline methods, GPT4-V and CLIP-Interrogator, and the best performing open-source IPRE with LLaVA-Next and LLaMA2, on this subset. The experimental results are presented in Table~\ref{tab:subset}.

Experimental results show that hand-created reverse prompts struggle to accurately reproduce the reference image. This highlights the challenges of manually designing reverse prompts and further emphasizes the importance of our method. The experimental results of commercial services shows that our ARPO method can achieve better performance than these commercial services, even though these commercial services are expensive. For the experimental results indicate that using closed-source models with vanilla prompt generation can achieve results similar to those using open-source models with enhanced prompt generation. Closed-source ARPO offers faster generation speeds, while open-source ARPO is completely free. Users are able to trade-off time and cost to choose the suitable method for their needs.

\begin{table}[H]
    \centering 
    \setlength\tabcolsep{9 pt}
    \begin{tabular}{ccccc}
\toprule
Method                    & CLIP-T & CLIP-I & DINO   & ViT   \\ \hline
Hand-crafted              & 25.07  & 72.88$\pm$0.04  & 40.69$\pm$0.84 & 37.10$\pm$0.06 \\
GPT4-V                    & 28.83  & 78.51$\pm$0.19  & 50.33$\pm$0.07  & 47.20$\pm$0.26 \\
CLIP-Interrogator         & 30.49  & 79.97$\pm$0.03  & 49.34$\pm$0.96  & 46.47$\pm$0.79 \\
ImageToPromptAI           & 29.20  & 78.34$\pm$0.16  & 48.92$\pm$0.15  & 45.54$\pm$0.17 \\
Phot.AI                   & 30.68  & 80.04$\pm$0.23  & 50.71$\pm$0.36  & 46.89$\pm$0.27 \\
Open-source ARPO          & 35.72  & \textbf{83.45$\pm$0.04}  & \textbf{56.36$\pm$0.65}  & 52.64$\pm$0.82 \\
Closed-source ARPO           & \textbf{36.01}  & 83.04$\pm$0.08  & 55.79$\pm$0.41 & \textbf{52.70$\pm$0.04} \\ \bottomrule
\end{tabular}
\caption{\textbf{Results of the image and prompt fidelity comparison on a subset.} Considering manpower resources and cost expenditures, we validate the performance of hand-crafted reverse prompt, commercial services, and closed-source ARPO methods on a 40 images subset.}
\label{tab:subset}
\end{table}

\subsection{The Results of AI-generated and Human-created Images}
We further provide the results of the PH2P, GPT4-V, CLIP-Interrogator and ARPO (LlaVA-Next, LlaMA2) on AI-generated and human-generated images so as to exclude the bias of combining the two types of images on the results. The results in Table~\ref{tab:seperate_result} demonstrate that our method achieves state-of-the-art performance on both two types of images.

\begin{table}[H]
    \centering 
\caption{Results for human-created and AI-generated images.}
\label{tab:seperate_result}
\resizebox{0.7\linewidth}{!}{
\begin{tabular}{lcccccccc}
\toprule
\multirow{2}{*}{Method} & \multicolumn{4}{c}{Human-created} & \multicolumn{4}{c}{AI-generated} \\ \cmidrule{2-5} \cmidrule{6-9}
& \texttt{CLIP-T} & \texttt{CLIP-I} & \texttt{DINO}  & \texttt{ViT} & \texttt{CLIP-T} & \texttt{CLIP-I} & \texttt{DINO}  & \texttt{ViT} \\ \hline
PH2P & 22.85 & 75.89 & 46.18 & 46.19 & 24.66 &76.64 & 46.36 & 38.62 \\
GPT-4V & 28.03 & 79.27 & 52.45 & 49.56 & 28.74 & 77.02 & 48.37 & 42.16 \\
CLIP-Interrogator & 29.85 & 79.27 & 49.17 & 48.57 & 31.27 & 81.85 & 51.71 & 44.03 \\
ARPO & \textbf{34.61} & \textbf{81.54} & \textbf{53.46} & \textbf{52.43} & \textbf{36.56} & \textbf{84.47} & \textbf{54.48} & \textbf{50.40} \\ 
\bottomrule
\end{tabular}
}
\end{table}

\subsection{The Results of Each Iteration}
\begin{figure}[ht]
    \centering
    \includegraphics[width=0.65\linewidth]{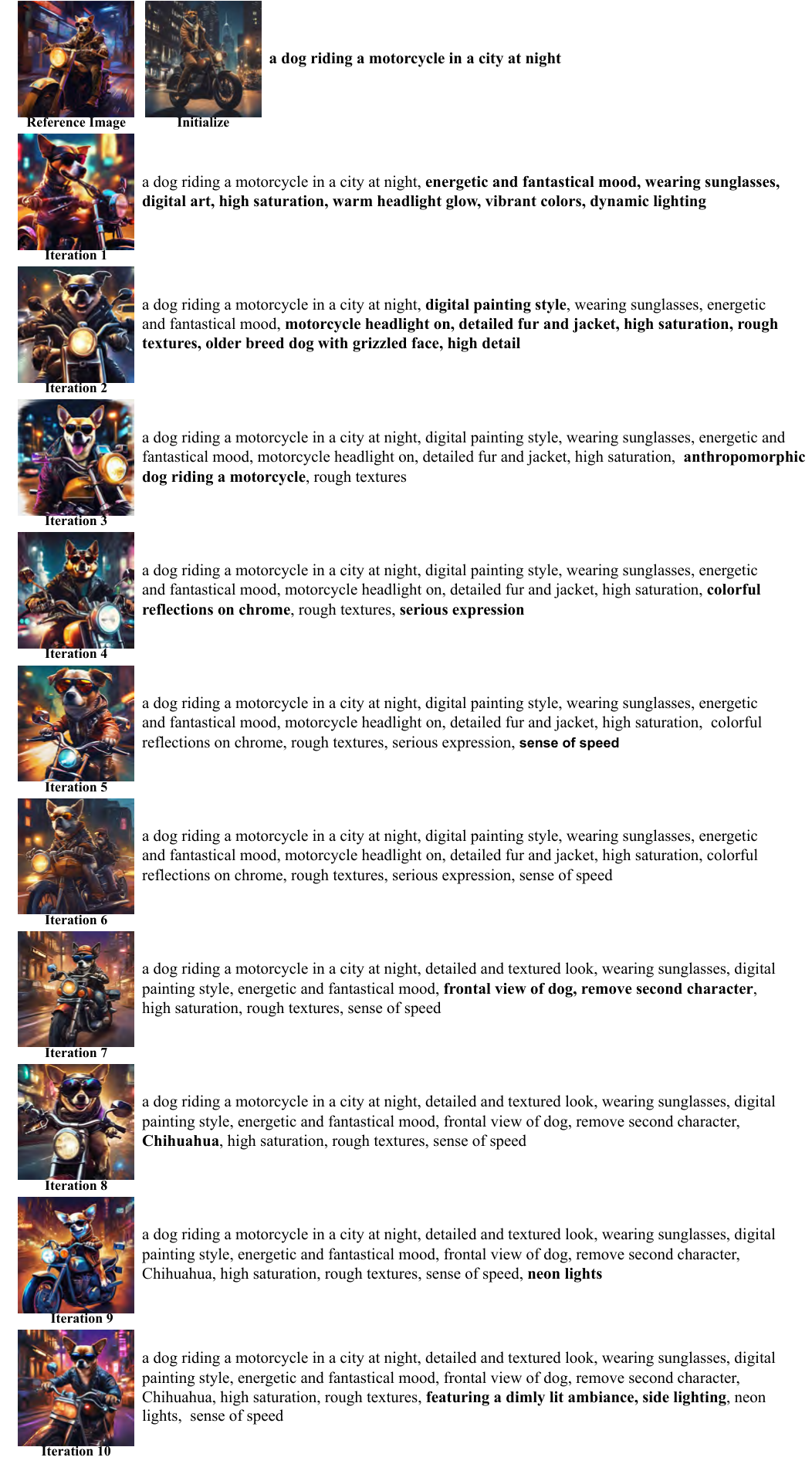}
    \caption{The reverse and generated image during the ARPO iterative process. We highlight the newly added prompts in each iteration by \textbf{bolding} them.}
    \label{fig:iterations_result}
\end{figure}

We present an example showcasing the reverse prompts and corresponding generated image during the ARPO iterative process. Figure~\ref{fig:iterations_result} illustrates the reverse prompt generated at each step of all iterations, alongside the images produced using these prompts. We highlight the newly added prompts in each iteration by bolding them. We observe that the new reverse prompts added during each iteration accurately address the discrepancies observed in the previously generated image. For example, in Iteration 3, the generated image features a smile dog, which does not match the atmosphere of the reference image. To rectify this,  "serious expression" is added to the reverse prompt in Iteration 4, which helps align the image more closely with the intended mood of the reference image. Furthermore, the generated image in Iteration 4 appears too static, "sense of speed" is added in the subsequent iteration to infuse dynamism into the image, thereby moving it closer to the aesthetics of the reference image.

\clearpage
\subsection{More Results of ARPO}
\begin{figure}[ht]
    \centering
    \includegraphics[width=0.85\linewidth]{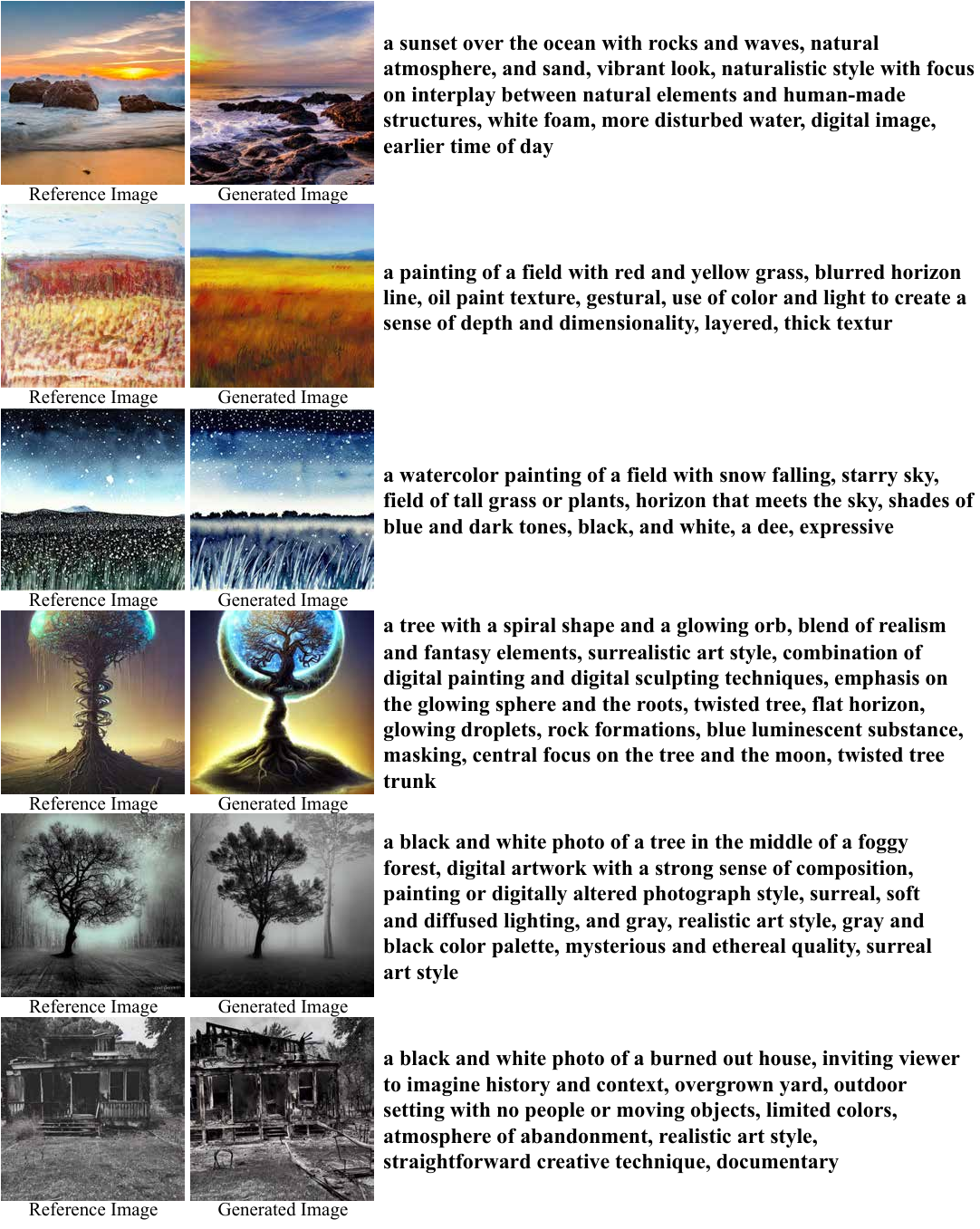}
    \caption{More results of ARPO.}
    \label{fig:IPR_result}
\end{figure}

We present additional examples demonstrating the capabilities of our ARPO method in reverse-engineering image prompts. Figure~\ref{fig:IPR_result} displays the reference image used for prompt reverse engineering, the reverse prompt generated by out ARPO method, and the image generated by the reverse prompt.

\clearpage
\subsection{Visualization Results and Corresponding Image Fidelity Results}
\label{appendix:visualization_and_fidelity}
\begin{figure}[ht]
    \centering
    \includegraphics[width=0.7\linewidth]{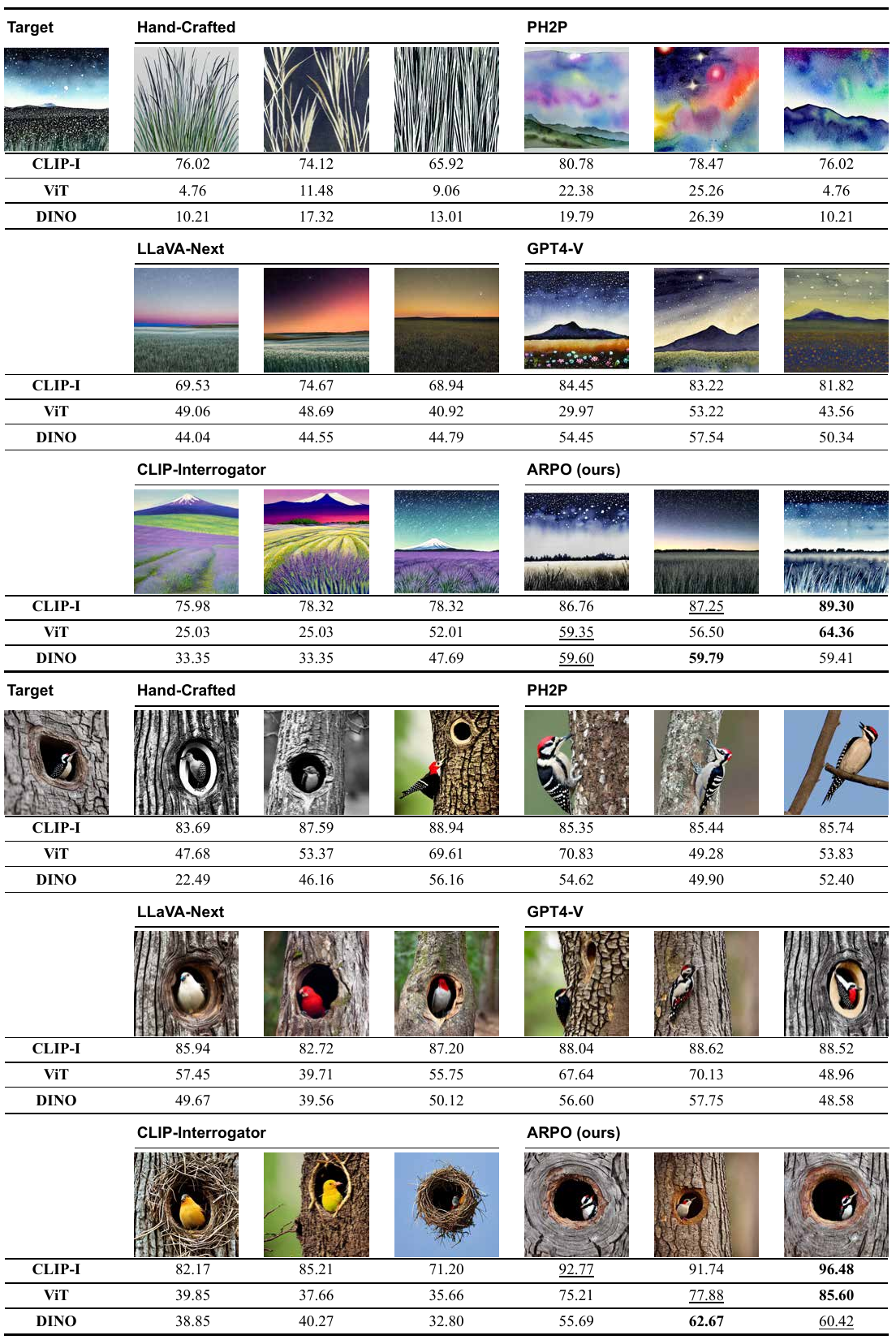}
    \caption{Visualization Results and Corresponding Image Fidelity Results.}
    \label{fig:visualization_fidelity}
\end{figure}

We provide the reconstructed images generated based on the reverse prompts and the image fidelity results between them and the reference images: \texttt{CLIP-I}, \texttt{ViT}, and \texttt{DINO}. According to Figure~\ref{fig:visualization_fidelity}, we observe a high correlation between the visualization results and the image fidelity results. Our method achieves superior results compared to other baseline methods in both qualitative and quantitative evaluations.

\clearpage
\subsection{More Results of the Comparison of Image Recreate}
\label{appendix:image_reconstruction}
\begin{figure}[ht]
    \centering
    \includegraphics[width=0.9\linewidth]{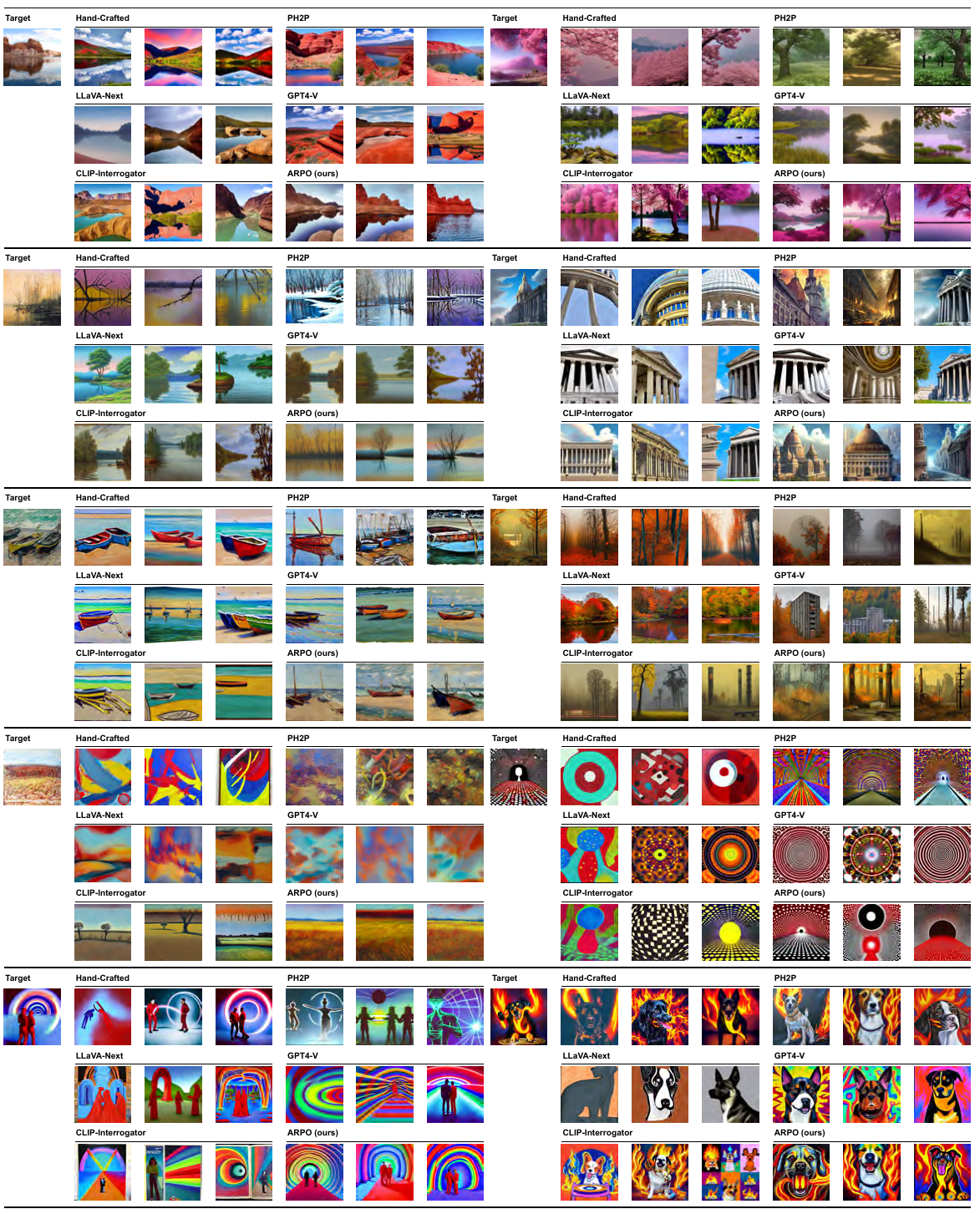}
    \caption{\textbf{Qualitative comparisons} of recreated images using the reverse prompts from different methods.}
    \label{fig:compare_appendix}
\end{figure}

We provide more comparison results of image recreate task in Figure~\ref{fig:compare_appendix}.

\clearpage
\subsection{Details of Novel Image Generation}
\label{appendix:novel_generation_details}
\begin{figure}[ht]
    \centering
    \includegraphics[width=0.83\linewidth]{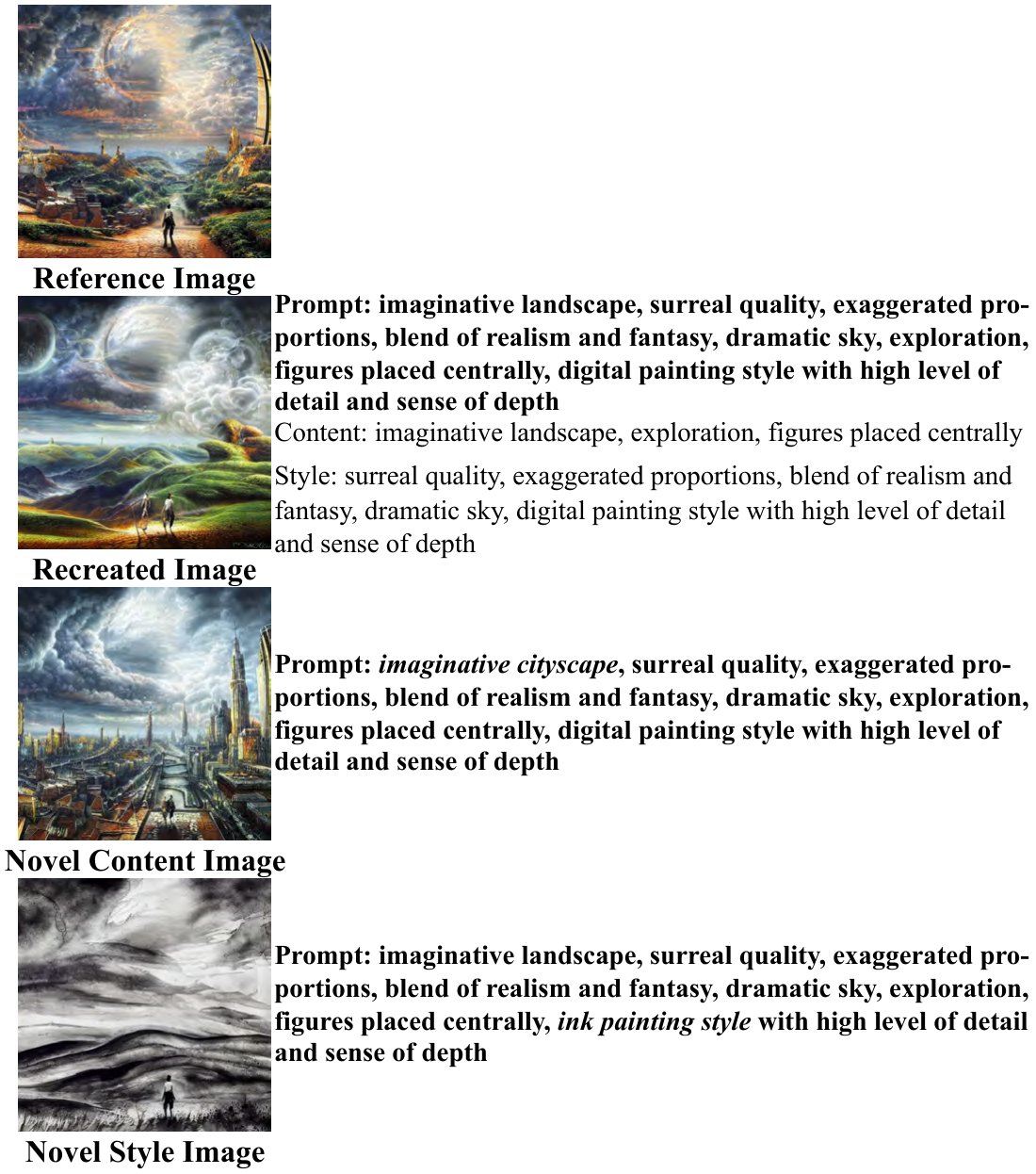}
    \caption{Novel images and their corresponding prompts. We use \textit{italics} to indicate the modifications made for generating novel images}
    \label{fig:control_method_details}
\end{figure}

We provide details on generating novel images based on reverse prompts in Section~\ref{sec:method:novel-image}. As shown in Figure~\ref{fig:control_method_details}, based on the reverence image, our APRO method can generate reverse prompts, which can then be used directly to create a similar recreated image. Since the reverse prompts we generate are in tag form, we can use LLMs to distinguish between content and style, assisting us in subsequent modifications. For the novel content image, we do not need to modify the style-related reverse prompts. In this example, we want to create an image of an urban scene, so we only need to replace "landscape" in the content-related prompts with "cityscape". Similarly, when we want to change the style of the image to ink painting, we only need to replace "digital painting" in the style-related reverse prompts.

\clearpage
\subsection{More Results of Novel Image Generation}
\label{appendix:image_modification}
We provide additional examples of new images generated by editing reverse prompts produced through the ARPO method. To facilitate user editing, our method additionally provides categorization of reverse prompts into content and style.  We edit the \textit{italicized} reverse prompt with the \textbf{bolded} words under the images to generated novel images.
\label{appendix:ner_creations}
\begin{figure}[ht]
    \centering
    \includegraphics[width=0.63\linewidth]{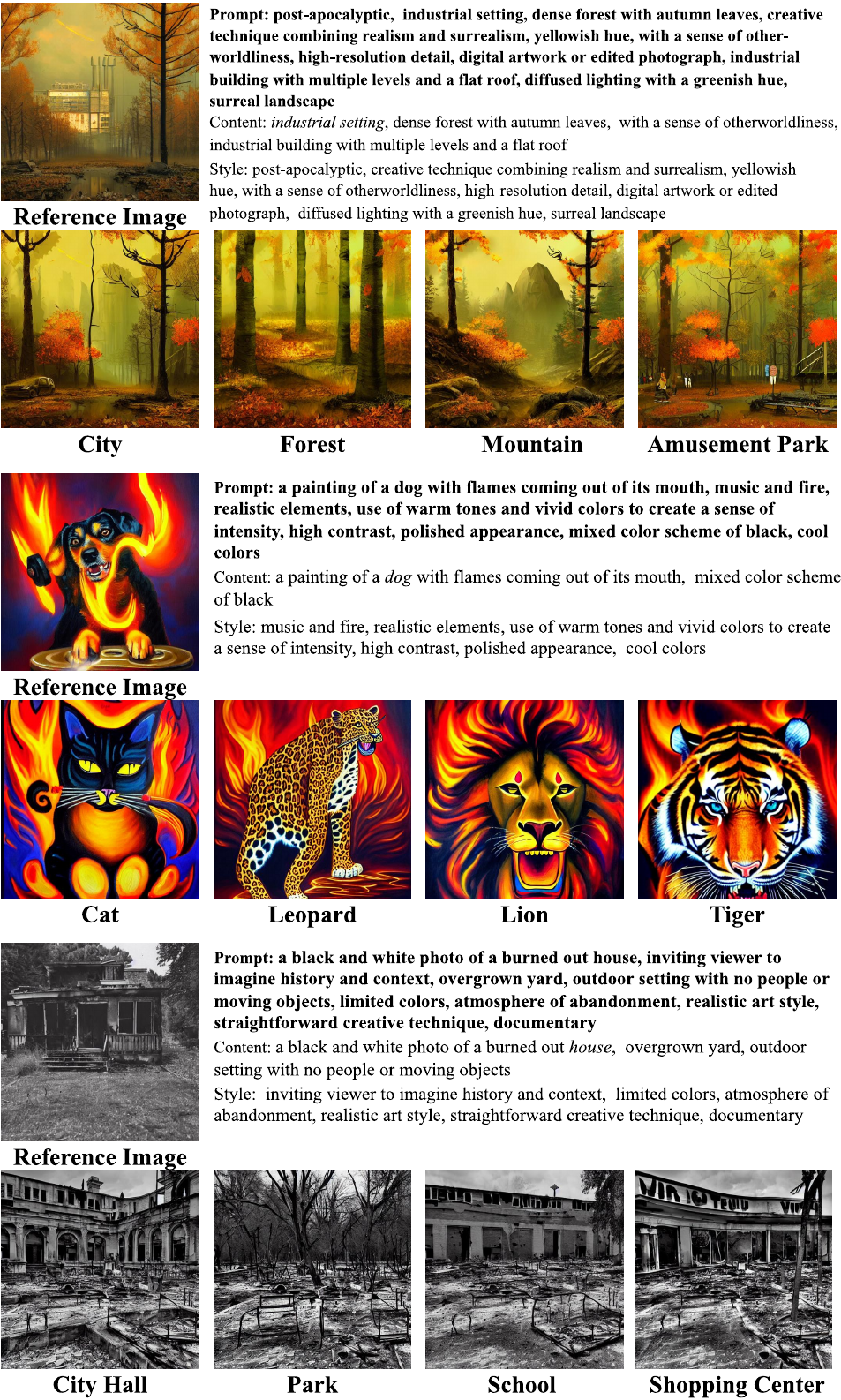}
    \caption{More results of Novel Image Generation. We replace the \textit{italicized} reverse prompt with the \textbf{bolded} words under the images to generated novel images.}
    \label{fig:control_results}
\end{figure}

\clearpage
\subsection{Comparison of Image Modification}
\label{appendix:comparison_image_modification}
We provide more comparison results of image modification. The comparison results are shown in Figure~\ref{fig:control_compare}. 
\begin{figure}[h]
    \centering
    \includegraphics[width=1\linewidth]{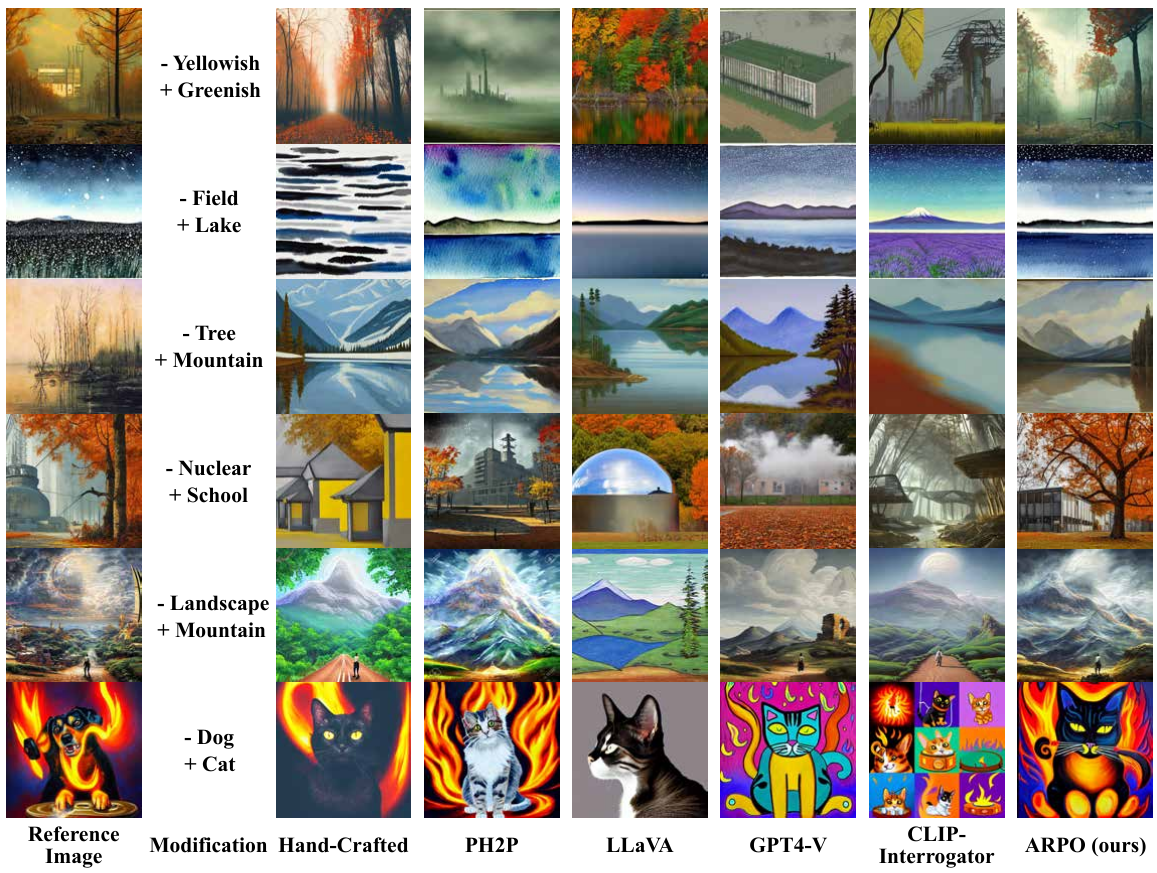}
    \caption{Comparison of editing reverse prompts for novel creations. For modification, "- Dog + Cat" indicates that we replace all instances of the word "dog" in reverse prompt with the word "cat".}
    \label{fig:control_compare_appendix}
\end{figure}

\clearpage
\subsection{Comparison of Image Fusion}
\label{appendix:image_fusion}
\begin{figure}[ht]
    \centering
    \includegraphics[width=1\linewidth]{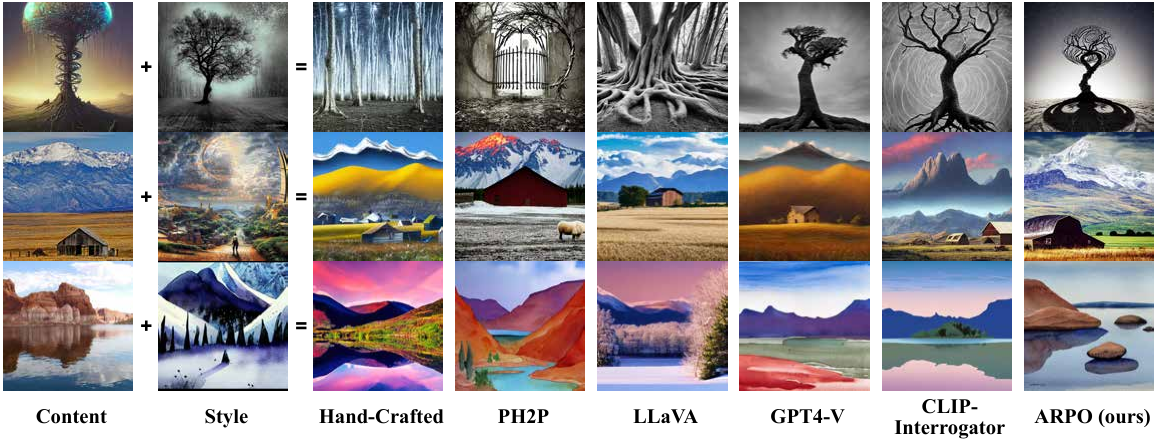}
    \caption{Comparison of image fusion. A new image is generated by merging style-related prompts from the "Style" image with content-related prompts from the "Content" image.}
    \label{fig:image_fusion_appendix}
\end{figure}

We provide more comparison results of image fusion. The results are shown in Figure~\ref{fig:image_fusion_appendix}.

\subsection{The Setting of User Study}
\label{appendix:user_study}
We conduct a user study to compare the image fidelity of our open-source and closed-source ARPO methods with hand-crafted reverse prompts, GPT4-V, CLIP-Interrogator, and commercial service (Phot.AI). We choose 40 cases for comparison. We ask users to rank the similarity between the generated image and the reference image based on content preservation, style approximation, and overall preference. Content Preference focuses on comparing the similarity of the image in aspects such as objects, characters, events, background, and actions. Style Preference focuses on comparing the similarity of the image in aspects such as artistic style, color, medium, and lighting. Overall Preference is a comprehensive comparison.

\begin{figure}[ht]
    \centering
    \includegraphics[width=0.83\linewidth]{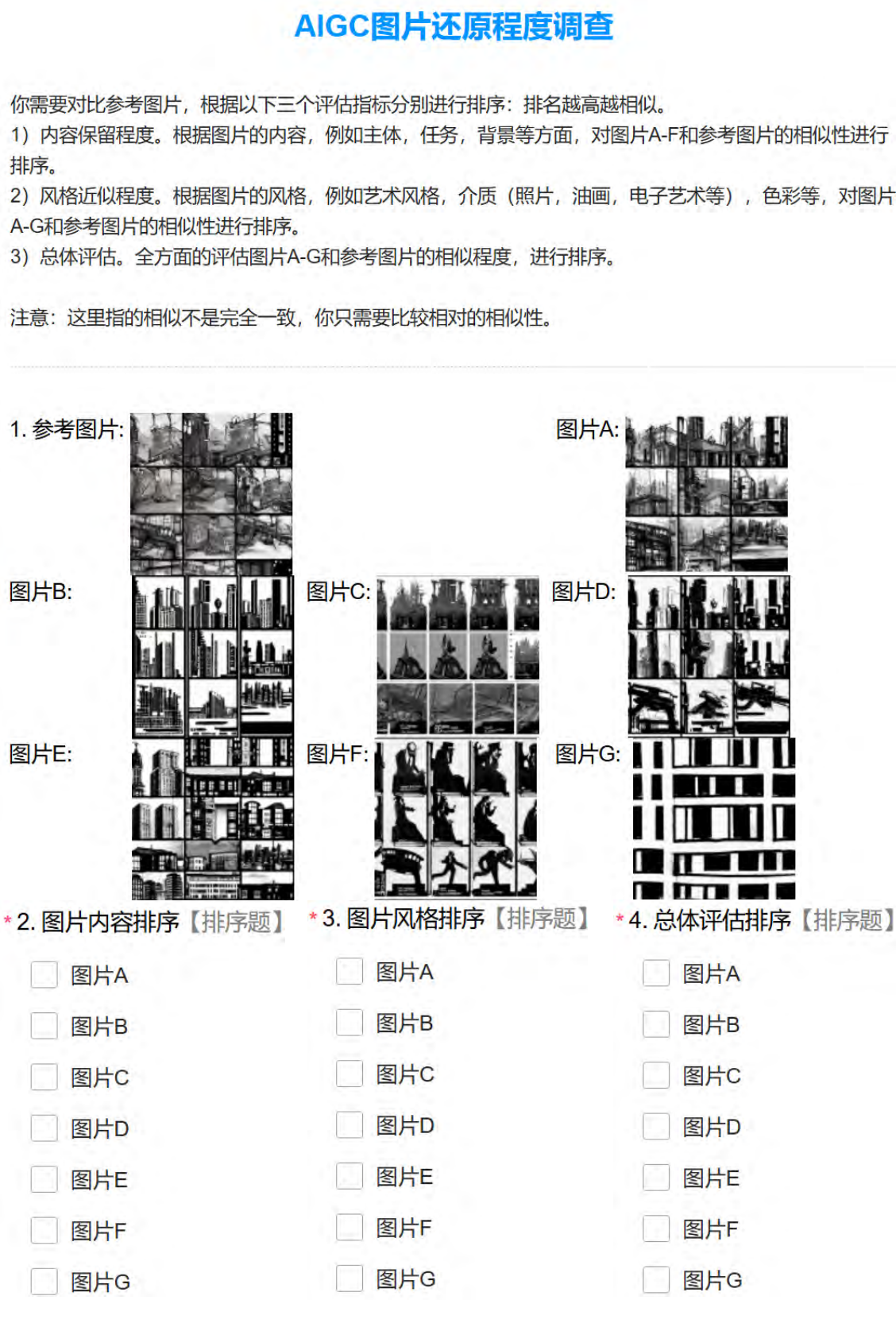}
    \caption{The screenshot of the questionnaire for user study}
    \label{fig:user_cn}
\end{figure}

\begin{figure}[ht]
    \centering
    \includegraphics[width=0.85\linewidth]{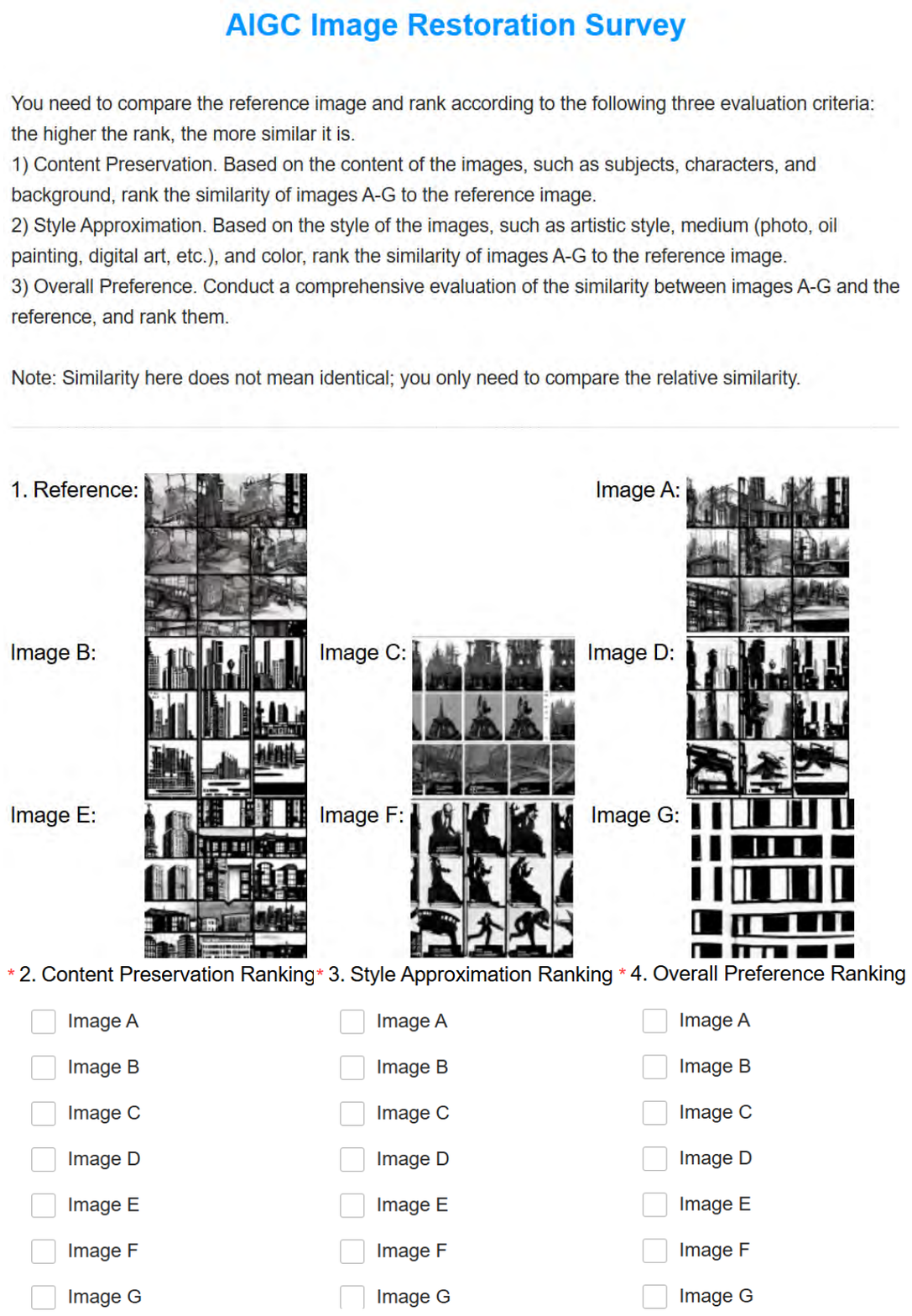}
    \caption{Translation of the screenshot of the questionnaire for user study}
    \label{fig:user_en}
\end{figure}

We collect feedback from 50 users. Considering portability and labor costs, we conduct this user study in China. We offer 4 CNY (about 0.55 USD) to each participant. Screenshot of the questionnaire and its English translation are shown in Figure~\ref{fig:user_cn} and Figure~\ref{fig:user_en}.

\end{document}